\documentclass[11pt]{article}

\usepackage[legalpaper, margin=1in]{geometry}

\usepackage[utf8]{inputenc} % allow utf-8 input
\usepackage[T1]{fontenc}    % use 8-bit T1 fonts
\usepackage{hyperref}       % hyperlinks
\usepackage{url}            % simple URL typesetting
\usepackage{booktabs}       % professional-quality tables
\usepackage{amsfonts}       % blackboard math symbols
\usepackage{nicefrac}       % compact symbols for 1/2, etc.
\usepackage{microtype}      % microtypography

\usepackage{amsfonts}       % blackboard math symbols
\usepackage{nicefrac}       % compact symbols for 1/2, etc.
\usepackage{microtype}      % microtypography
\usepackage{subfigure} 
\usepackage{graphicx} % more modern
\usepackage{enumitem}

\usepackage{bm}
\usepackage{amsmath}
\usepackage{amsthm}
\usepackage{pdfpages}
\usepackage[numbers]{natbib}

\renewcommand{\b}{\mathbf}
\newcommand{\bt}[1]{\mathbf {\tilde {{#1}}}}
\newcommand{\bb}[1]{\bar {\mathbf {{#1}}}}
\newcommand{\bh}[1]{\hat {\mathbf {{#1}}}}
\newcommand{\dist}[1]{\text{dist} \left( {#1} \right)}
\newcommand{\tr}[1]{\text{tr} \left( {#1} \right)}
\newcommand{\Ln}{\left \Vert}
\newcommand{\Rn}{\right \Vert}
\newcommand{\E}[1]{\mathbb E \left[ {#1} \right]}
\renewcommand{\P}[1]{\mathbb P \left[ {#1} \right]}
\newcommand{\Var}[1]{\textrm{Var} \left[ {#1} \right]}

\newcommand{\Pgiven}[2]{\mathbb P \left[ {#1} \ \middle| \ {#2} \right]}

\newtheorem{theorem}{Theorem}
\newtheorem{lemma}{Lemma}

\title{Spectral Algorithm for Low-rank Multitask Regression}
\author{Yotam Gigi\footnotemark[1]  \quad
        Ami Wiesel\footnotemark[1]  \quad
        Sella Nevo\footnotemark[2]  \quad
        Gal Elidan\footnotemark[1] \vspace{0.1cm} \\ 
        Avinatan Hassidim\footnotemark[3]  \quad
        Yossi Matias\footnotemark[4]}
\date{September 2019}

\begin{document}

\maketitle

\begin{abstract}
Multitask learning, i.e. taking advantage of the relatedness of individual tasks in order to improve performance on \emph{all} of them, is a core challenge in the field of machine learning. 
We  focus on matrix regression tasks where the rank of the weight matrix is constrained to reduce sample complexity. We introduce the common mechanism regression (CMR) model which assumes a shared left low-rank component across all tasks, but allows an individual per-task right low-rank component. This  dramatically  reduces  the number of samples needed for accurate estimation.
The problem of jointly recovering the common and the local components has a non-convex bi-linear structure. We overcome this hurdle and provide a provably beneficial non-iterative spectral algorithm.
Appealingly, the solution has favorable behavior as a function of the number of related tasks and \emph{the small} number of samples available for each one. 
We demonstrate the efficacy of our approach for the challenging task of remote river discharge estimation across multiple river sites, where data for each task is naturally scarce. In this scenario sharing a low-rank component between the tasks translates to a shared spectral reflection of the water, which is a true underlying physical model. We also show the benefit of the approach on the markedly different setting of image classification where the common component can be interpreted as the shared convolution filters.
\end{abstract}

\renewcommand{\thefootnote}{\fnsymbol{footnote}}
\footnotetext[1]{Google Research and The Hebrew University}
\footnotetext[2]{Google Research}
\footnotetext[3]{Google Research and Bar-Ilan University}
\footnotetext[4]{Google Research and Tel-Aviv University}

\section{Introduction}

% One sentence: Sometimes there is not enough data for learning, for example floods
Powerful machine learning models in general, and deep models in particular, can provide state of the art performance in a wide range of settings, but require large amounts of data to train on. In many applications, however, labeled data for a particular task can be scarce. For example, we are motivated by the important problem of estimating river discharge (water volume per second) using remote sensing, namely multispectral satellite imagery \cite{smith1997satellite}. For a particular location of interest, e.g. in a flood-prone section of a river, the number of available images and corresponding gauge measurements for training can be quite small since satellite images of the same location are taken at low frequency.

% One sentence: But, sometimes we have a lot of related such tasks, for example floods.
But, it is also often the case that we have access to many related problems. Continuing the above example, the multiple problems of predicting water discharge at different river locations are all related via the physics underlying the reflection of the satellite signal as it hits the water. Intuitively, we should be able to leverage this relatedness to learn better predictive models for \emph{all} problems since each local measurement gives us a "clue" as to the nature of the common physical mechanism. This general setting has a long history in machine learning: inductive transfer, transfer learning, and multitask learning are all closely related variants of the framework (see, e.g. \cite{Thrun:1996, InductiveTransfer:1997, Caruana:1997, Baxter:2000, Pan:2010}).

% One sentence: The CMR model, intuition in remote sensing and in multiclass conv net
In this work, we introduce \emph{common mechanism regression} (CMR), a low-rank matrix recovery model with a shared component in one of its dimensions and decoupled components in the other. Intuitively, the common part acts as a joint feature selection mechanism \cite{zhang2017survey}, which allows for a much simpler local part. The CMR model is motivated by remote sensing, where the common dimension corresponds to the spectral reflection mechanism which is shared across locations, and the decoupled dimension is associated with the independent spatial tasks \cite{yuan2017spectral}. Another motivation for the model is in the context of modern convolution networks for multiclass classification, where the common dimension corresponds to filters which are common for all classes, and the decoupled dimension is associated with the last layer which is disjoint across classes. 

% One sentence: CMR is complex with low number of d.o.f.
In the context of remote river discharge estimation, CMR translates directly to the following intuitive claim: a per-site learning algorithm does not need the whole multi-spectral image of the river in order to estimate the discharge, but only few global linear combinations of its spectral bands. In the context of multi-label classification using convolutional networks, CMR similarly suggests that a few down-sampled convolution masks are sufficient for the per-label classification, instead of the whole image.
We show by extensive experiments that the CMR model has the complexity needed for the above problems, while maintaining a particularly low number of degrees of freedom ($d.o.f.$) which makes it especially suitable for data scarce problems.% like river discharge estimation.

% One sentence: Non convex, thus we present the CMR algorithm
%But, to enjoy the benefits of the CMR model and recover its parameters, we must solve a non-convex bi-linear optimization problem. We take our inspiration from initialization techniques used in the context of low-rank optimization problems \cite{candes2015phase, netrapalli2013phase,tu2015low,vaswani2017low, huang2006extreme} and introduce the CMR \emph{algorithm}, a simple closed form spectral algorithm for recovery of the model parameters in the non-convex scenario. Our theoretical analysis shows that, under technical statistical conditions that allow for realistic and correlated features, the algorithm can recover the model parameters with a number of samples that matches the number of degrees of freedom of the model, up to a multiplicative constant.

% One sentence: Non convex, thus we present the CMR algorithm
A key point in CMR is that it recovers the common part (i.e., the global linear combinations) and the local part (i.e., the per-task regressor) jointly, which leads to a non-convex bi-linear optimization problem. We take our inspiration from initialization techniques used in the context of low-rank optimization problems \cite{candes2015phase, netrapalli2013phase,tu2015low,vaswani2017low, huang2006extreme} and introduce the CMR \emph{algorithm}, a simple closed form spectral algorithm for recovery of the model parameters in the non-convex scenario. Our theoretical analysis shows that, under technical statistical conditions that allow for realistic and correlated features, the algorithm can recover the model parameters with a number of samples that matches the number of degrees of freedom of the model, up to a multiplicative constant.

\begin{figure*}[t]\label{patna_fig}
\begin{center}
\begin{tabular}{cc}
\includegraphics[height=0.23\textwidth]{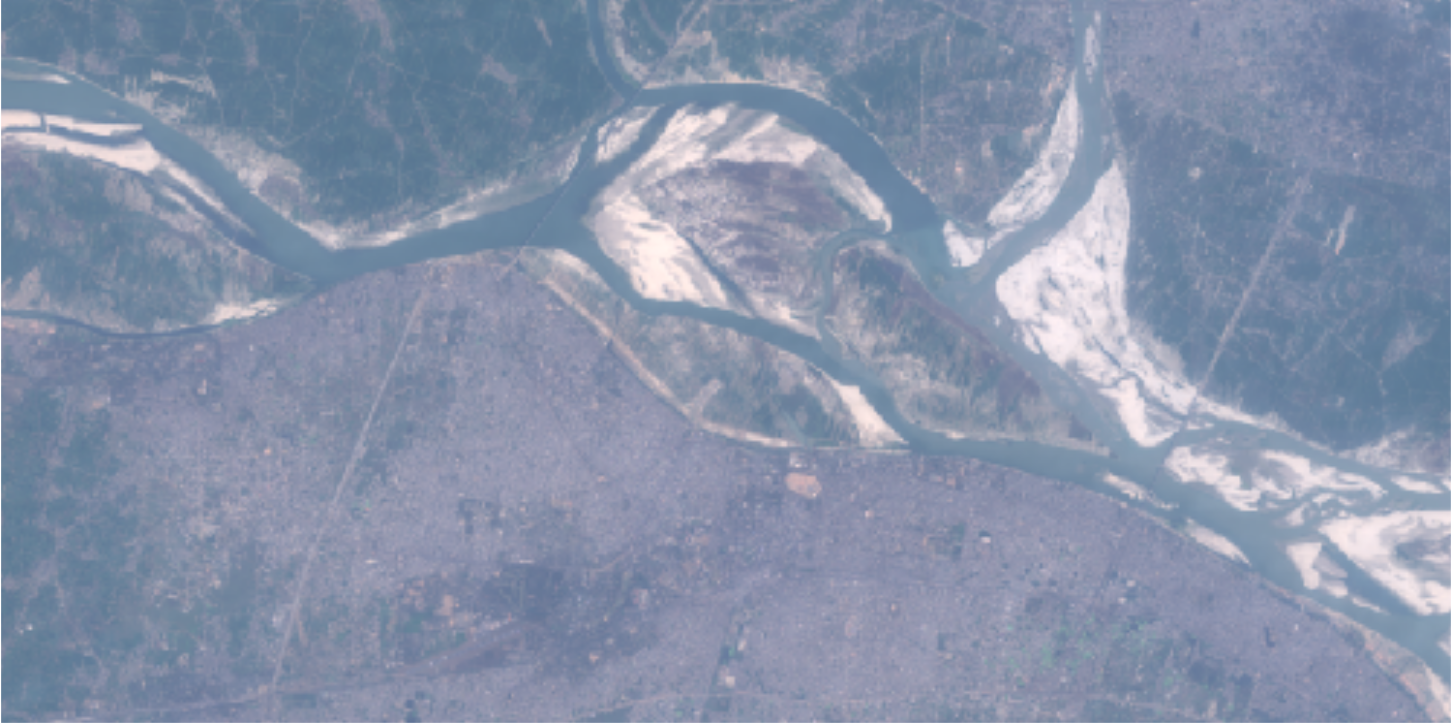} &
\includegraphics[height=0.23\textwidth]{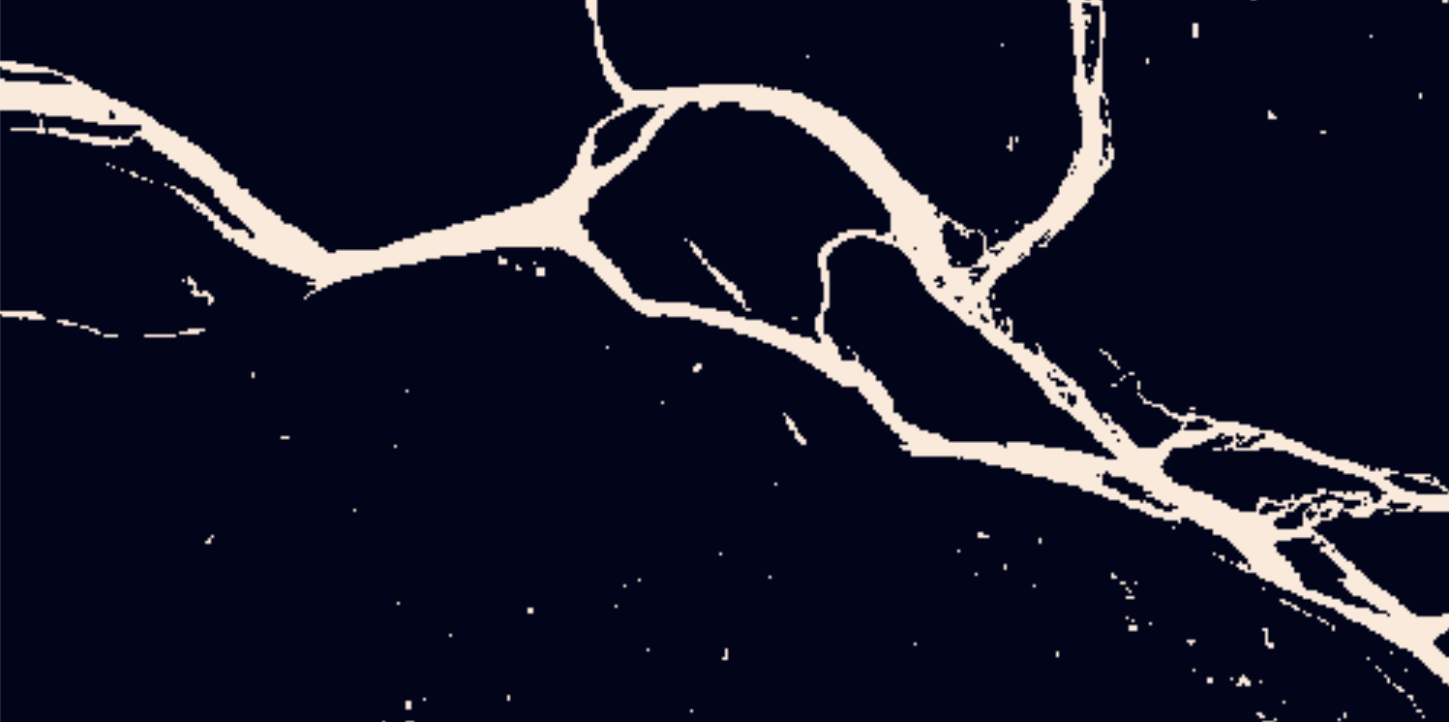}
\end{tabular}
\caption{\small{An RGB satellite image of a portion of the Ganges river in India (left), along with its thresholded CMR approximation ($R=1$) that accurately identifies the water pixels (right). The generated $\b W$ parameters are $[ -0.4,   0.1,   1.3,
        -6.9,   4.1,   4.1,
        -1.6,  -0.6,   0.4, 0.0, 0.0]$ and correspond to the 11 spectral bands of the satellite LANDSAT8.}}
\end{center}
\vspace{-0.5cm}
\end{figure*}

% One sentece: We test CMR on floods dataset and it works
We demonstrate the efficacy of our approach by applying CMR to the two markedly different settings described above. In the first scenario, the goal is to perform remote river discharge estimation across multiple river sites using satellite imagery. 
Specifically, we use multi-spectral images from the LANDSAT8 mission \cite{roy2014landsat} and ground truth discharge labels from the United States Geological Survey (USGS) website. The results are quite convincing and are exemplified in Figure \ref{patna_fig}, which demonstrates the ability of CMR to accurately identify the water pixels within a flood prone region of the Ganges river in India (see Section \ref{sec:experiments} for more details).

\begin{figure*}[t]
\centering
\includegraphics[trim=0 0.2cm 0 0.9cm, clip, height=0.28\textwidth]{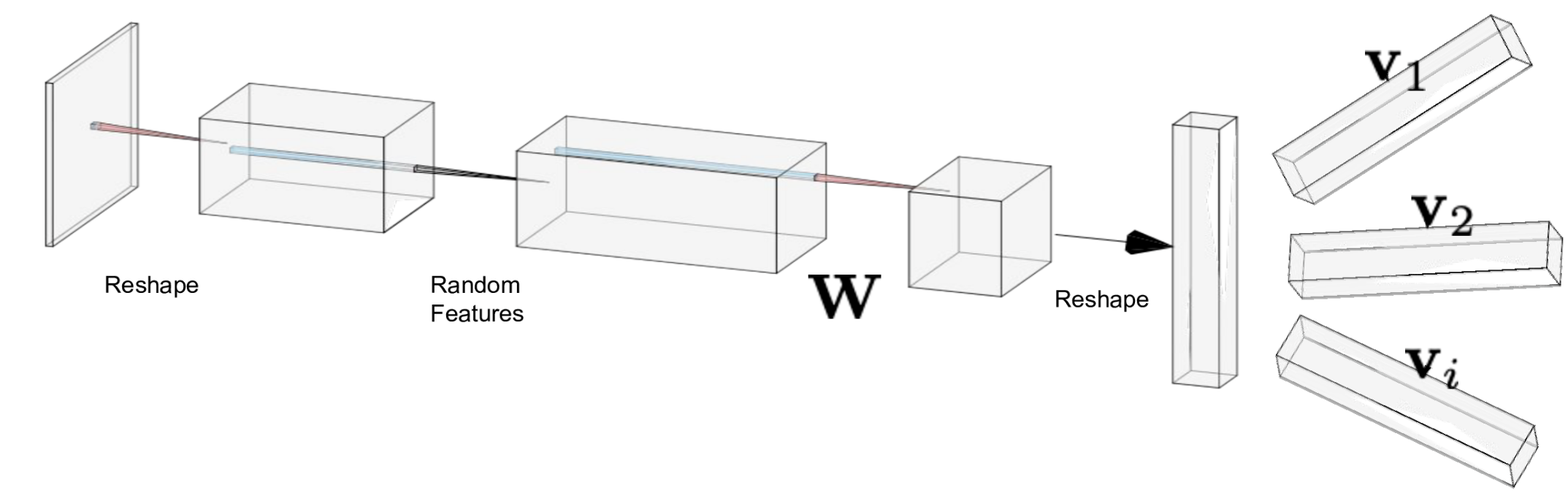}
\vspace{-0.3cm}
\caption{\small{Illustration of CMR usage in image multi-class classification: the first step is a convolution-like reshaping that transforms an image to a down-sampled version with multiple channels (bands). Next, the bands are up-lifted to higher dimensions via random non-linear mappings. A common mechanism is then applied in the bands dimension to construct the important features. Finally, individual regressions are performed per binary classification task. Graphics were generated via http://alexlenail.me/NN-SVG/}}
\label{cmr_classification_drawing}
\end{figure*}

The second scenario we consider is image classification using modern convolution networks. The first layers in such networks transform an RGB image into a down-sampled tensor with multiple channels. Here, CMR applies a common mechanism across the channels, followed by a per class simple linear spatial regression, as illustrated in Figure \ref{cmr_classification_drawing}. The advantage of CMR using the MNIST \cite{lecun1998gradient} and the Street View House Number (SVHN) \cite{netzer2011reading} datasets is clearly demonstrated.

\subsection{Notations}
%%%%%%%%%%%%%%%%%%%%%%
We use bold capital letters for matrices and bold lowercase letters for vectors. We use  $\Ln \b v \Rn$  to indicate the standard $\ell_2$ euclidean norm, and $\Ln \b v \Rn_p$ to indicate a specific $p$-norm. We use $\tr{\b A}$ to indicate the trace of the matrix $\b A$, and $\b A^\dag$ to denote its pseudo inverse. For matrix norm, we use $\Ln \cdot \Rn_2$ to indicate the matrix operator norm for $\ell_2$, and $\Ln \cdot \Rn_F$ to indicate the matrix Frobenious norm.

In addition, we use use $\b A \otimes \b B$ to indicate a Kronecker product between the matrix $\b A$ and $\b B$, and we denote by ${\rm{vec}}(\b A)$ the vector obtained by stacking the columns of $\b A$ one over the other. We use $\mathbb E[\b x]$ to indicate the expectation of a random variable $\b x$ and $\text{cov}(\b x)$ its covariance matrix. We use the term $\b {eigv}_R(\b A)$ to indicate the matrix whose columns are the $R$ eigenvectors of $\b A$ corresponding to the largest eigenvalues. To measure the distance between two subspaces spanned by the columns of the orthogonal matrices $\b U$ and $\b V$, we use a 2-projection norm defined as 
$\dist{\b U, \b V}=\Ln \b U \b U^\dag -  \b V \b V^\dag  \Rn_2 $
which is equivalent to the largest principal angle between them.

\section{Related work}
{\bf{Low-rank Matrix Optimization}}: There is a large body of literature on recovering a low-rank matrix given partial or noisy observations. These include matrix completion works and phase retrieval problems \cite{candes2015phase, netrapalli2013phase,recht2010guaranteed, zhu2017global}, alongside many others. The problems are non-convex but there are well understood conditions for successful recovery, as well as efficient algorithms that attain them. The CMR model can also be formulated as a reduced rank matrix recovery problem, but this formulation involves structured and correlated sensing matrices that do not fit into existing methods and theory. Instead, our work generalizes the spectral initialization proposed in \cite{candes2015phase, netrapalli2013phase,tu2015low,vaswani2017low} along two axes. First, CMR introduces a common mechanism over multiple tasks. Second, CMR uses a whitening pre/post-processing stage that allows real-world correlated features. We present the theoretical implications and demonstrate the empirical advantages of both extensions.

{\bf{Matrix Variate Normal}}: The matrix variate normal density, also know as the Kronecker model, is a matrix-valued probability distribution that allows for structured correlation between the matrix elements \cite{dutilleul1999mle, werner2008estimation, wiesel2012geodesic}. This model is commonly used in the context of multitask learning where, as detailed in \cite{zhang2010learning}, it characterizes both task relatedness and feature representation \cite{stegle2011efficient,bonilla2008multi}.
In the context of low-rank models, previous works typically assume engineered sensing structures, e.g., independent identically distributed ($i.i.d.$) sensing vectors/matrices \cite{recht2010guaranteed, netrapalli2013phase, candes2015phase, vaswani2017low, tu2015low, huang2019solving}. In contrast, in CMR we cope with realistic correlated features. CMR generalizes previous spectral initialization schemes (as well as their theoretical analyses) to the matrix-variate normal case.

{\bf{Multitask Learning}}: The CMR algorithm solves several tasks jointly in order to improve performance. Thus, it is natural to compare it to other multitask learning (MTL) algorithms. 
The body of literature on MTL is quite large, and is commonly classified into several main approaches \cite{zhang2017survey, zhang2018learningto}. CMR is mostly inspired by the low-rank MTL approach, and specifically by the seminal work of \cite{ando2005framework}. Similarly to CMR, these works utilize a shared low-rank linear feature subspace in order to improve performance.
The CMR algorithm can also be viewed as a feature learning MTL \cite{obozinski2006multi,liu2009multi} variant, where the common mechanism defines jointly-learned features to be used in each individual task.
%Apart from the low-rank approach, CMR uses similar ideas from other feature learning MTL approaches \cite{obozinski2006multi,liu2009multi} since they all try to jointly learn better features for the tasks.
Uniquely, CMR considers a variant of these ideas in which features can be naturally organized along two axes (e.g. a label per hyper-spectral image which has both spatial and spectral axes), where the common subspace operates only on one of the dimensions. Importantly, we show that in this case, there is a closed form spectral algorithm with theoretical guarantees.

{\bf{Random Features}}: Practically, CMR can be used as a natural extension to extreme learning machines or regression with random features \cite{huang2006extreme,rahimi2009weighted}. These methods provide non-linear capabilities by random non-linear lifting to higher dimensions and only optimize the final linear regression layer. In contrast, CMR optimizes the last two layers (see Figure \ref{cmr_classification_drawing}), exploiting the power of many tasks in order to achieve lifting to a much higher dimension, with a limited number of samples per task. This view of CMR makes it clear that CMR is naturally applicable to modern convolution networks where an RGB image is down sampled into a lower resolution image with more channels. Concretely, we propose to use a large number of random channels and let CMR automatically choose a common subspace within them. See Section \ref{sec:experiments} for a demonstration of this use case.

\section{Common Mechanism Regression (CMR)}

Our model consists of $I$ independent regression tasks. For simplicity, we assume that each regression has exactly $T$ pairs of labels and features, i.e.
\begin{eqnarray} \nonumber
    \left\{y_{it},\b{X}_{it}\right\}_{t=1}^T \quad i=1,\cdots,I
\end{eqnarray}
where $y_{it}$ are scalar labels, and $\b{X}_{it} \in \mathbb R^{B \times P}$ are the features. The CMR model addresses problems where the features are matrix-valued and inherently aligned along two axes. Two motivating applications are:
\begin{itemize}
    
    \item {\bf{Remote river discharge estimation}} where there are $I$ different river locations, each with $T$ temporal observations. In this context, an observation is a multi-spectral image of the river location consisting of $B$ spectral bands and $P$ pixels, which are the two axes along which features are aligned \footnote{In practice, the observations are typically a tensor with $B$ channels of $\sqrt{P}\times \sqrt{P}$ pixels. For notation purposes, we flatten the images into vectors of length $P$ and work with $B\times P$ observation matrices.}. Every such observation is accompanied by a matching scalar label representing the river discharge.
    
    \item {\bf{Multilabel classification using convolution networks}} where every image sample passes through convolution-like reshaping, and thus has two distinct axes: $P$, the specific image patch, and $B$, the specific convolution channel. In addition, every per-label classification can be treated as a separate task. Thus, there are $I$ different per-label tasks, each with $T$ samples.
    
\end{itemize}

% One sentence: The CMR model (y = tr W'XV) and the rank R, with the flood and conv intuition
We introduce the \emph{Common Mechanism Regression (CMR)} as a natural multitask model for this matrix-valued structure. CMR is bilinear model, where the first component is common and parameterized by a matrix $\b W \in \mathbb R^{B \times R}$, followed by a decoupled per-task parameter $\b V_i \in \mathbb R^{P \times R}$:
\begin{equation}
\begin{gathered}
    \label{model}  y_{it} = \tr{\b W^T\b X_{it} \b V_i}%  \equiv \tr{\b X_{it} \b S_{i}}  \\
    %\b S_i \equiv \b V_i \b W^T
\end{gathered}
\end{equation}
The common $\b W$ reduces an observation of dimension $B\times P$ to a much smaller $R \times P$. 
%
%In the special case of $R=1$ and in the context of river-discharge estimation, this implies that the discharge depends linearly on a monochrome image which is a linear combination of the spectral bands, where this combination is shared across the river sites.
%
In the context of river-discharge estimation and in the special case of $R=1$, this implies that the discharge depends linearly on a monochrome image which is a linear combination of the spectral bands. The model further suggests that this combination is common to all the river sites.
In the context of multi-label classification based on convolutional networks, $\b W$ is a set of $R$ convolutional masks shared between all the per-label classification tasks.
%translates directly to finding the best linear combination of the spectral bands that will allow the best per-site linear regression on the pixels.

% One sentence: CMR model as shared low rank model
A close look at the CMR model also reveals that it is a generalization of reduced rank models. Specifically, a low-rank component is shared across many low rank matrix recovery tasks. This can be seen by defining $\b S_i \equiv \b V_i \b W^T$ and thus eq. (\ref{model}) becomes
\begin{equation} \nonumber
y_{it} = \tr{\b X_{it} \b S_{i}}
\end{equation}
where the matrices $\b S_{i}$ are rank $R$ matrices that all share a common right subspace. 

% One sentence: The sample complexity saved by the CMR model vs low rank and full regression
The advantage of the CMR model is most evident when looking at the number of degrees of freedom (d.o.f.) per individual task. Without the rank requirement, we get a standard per-task linear model where the matrices $\b S_i$ from eq. (\ref{model}) are full rank, thus there are $B\cdot P$ d.o.f. per task. When a CMR-like bi-linear model is used where $\b W$ is not shared between the tasks, i.e. $y_{it} = \tr{\b W_i \b X_{it} \b V_i}$, the number reduces to $R(B+P)$. Finally, and when $\b W$ is shared across the tasks as presented in eq. (\ref{model}), we achieve a further substantial reduction to $R(P+B/I)$ d.o.f.
%where this is most apparent when $B \gg P$ \ynote{talk with Ami - this assumption is not very good for us :)}

% One sentence: The optimization problem
Having defined the CMR model, our goal is to recover the common parameter $\b W$ and, if possible, to also identify the local $ \b V_i $'s. Concretely, the CMR \emph{optimization} problem is defined as follows:
\begin{align} \begin{split} \label{wv_opt_problem}
\min_{\bb W, \bb V_1, ... \bb V_I}
    &\sum_{i,t} \left( y_{it} - \tr{\bb  W^T \b X_{it} \bb V_i} \right)^2 .
\end{split}\end{align}
% One sentence: The CMR problem as low rank problem
Note that the overall structure is linear in the features, but has a bilinear parameterization, and thus the recovery of $\b W$ and $\b V_i$ in eq. (\ref{model}) is not straightforward. 
In the context of standard reduced rank matrix recovery, it has been shown that local minima can be avoided by a spectral initialization algorithm \cite{tu2015low, candes2015phase, netrapalli2013phase,vaswani2017low}. We take our inspiration from such methods and introduce the CMR \emph{algorithm}, a generalization of existing techniques to the multitask low-rank regression scenario.

\begin{figure*}[b]
\vspace{0.2cm}
\begin{center}
    \bf{Algorithm 1: CMR} \\
    \vspace{1mm}
    \begin{tabular}{p{11cm}}
    \hline 
    \vskip -0.3cm
    %\hline
    {\begin{align}
        &\emph{Cross correlation}:   && \textstyle{\b Z_i \leftarrow \frac 1T \sum _t y_{it} \b X_{it}, \quad \quad i = 1,\cdots,I }\nonumber\\
        &\emph{Outer product}:       && \textstyle{\bh A \leftarrow \frac 1I \sum_{i} \b Z_i \b Z_i ^T}\nonumber  \\
        &\emph{Cov estimation}:      && \textstyle{\bh \Gamma \leftarrow \frac 1{ITP} \sum_i \sum_t \b X_{it} \b X_{it}^T } \nonumber \\
        &\emph{Pre-subspace whitening}:           && \bh B \leftarrow \bh\Gamma ^ {-1/2}\bh A \bh \Gamma ^ {-1/2}\nonumber  \\
        &\emph{Subspace}:            && \bh W' \leftarrow  \b{eigv}_R \left( \bh B \right)\nonumber   \\
        &\emph{Post-subspace whitening}:        && \bh W \leftarrow \bh \Gamma ^{-1/2} \bh W' \nonumber \\
        &\emph{Regressions}:         && \bh V_i \leftarrow {\rm{Ridge}}(\bh W^T\b X_{it},\b y_{it},\alpha),\nonumber 
                                        \quad i = 1,\cdots,I\nonumber
    \end{align} }\\
    \end{tabular}
\end{center}
\vspace{-1.0cm}
\end{figure*}

% One sentence: Some intuition on the algorithm.
%The CMR algorithm is outlined in Algorithm 1 below. To gain some intuition, consider first the \emph{cross correlation} step at the beginning and \emph{subspace} stage that appears later. The first computes a sample cross correlation between the labels $y_{it}$ and the features $\b X_{it}$ which, under $i.i.d$ conditions (i.e., identity covariance), should converge to $\b W\b V_i^T$. Then, the leading $R$ dimensional left-subspace are identified in the subspace phase. Diving deeper into the details, note that the CMR algorithm is only expected to recover the shared $\b W$ and not the individual $\b V_i$'s. Thus, we take another outer product and average over the $I$ tasks to get one $B \times B$ matrix $\bh A$. With enough $i.i.d$ features, the leading subspace of this matrix should recover $\b W$. More generally, $\bh B$, which is a whitened version of $\bh A$, holds all the information needed to construct $\bh W$, via its eigenspace.

% One sentence: Some intuition on the algorithm.
The CMR algorithm is outlined in Algorithm 1 below. To gain some intuition, consider $i.i.d.$ conditions where it holds that that $\bh \Gamma = \b I$ and $\bh A = \bh B$. 
%In this case, it can be seen that the \emph{cross correlation} phase computes a per-task cross correlation between the labels $y_{it}$ and the features $\b X_{it}$, where standard statistical analysis shows that this cross correlation converges to $\b W\b V_i^T$. Since we are interested in recovering the common mechanism $\b W$ and not the individual $\b V_i$, we take a per-task outer product and average over the $I$ tasks to get one $B \times B$ matrix $\bh A$, where its leading $R$ dimensional eigen-subspace converge to $\b W$. Thus, \emph{subspace} phase identifies this subspace. 
In this case, the algorithm starts by computing a per-task cross correlation between the labels $y_{it}$ and the features $\b X_{it}$ (see phase \emph{cross correlation} in the algorithm). Standard statistical analysis shows that this cross correlation converges to $\b W\b V_i^T$.
Since we are interested in recovering the common mechanism $\b W$ and not the individual $\b V_i$, we take a per-task outer product and average over the $I$ tasks to get one $B \times B$ matrix $\bh A$.
The algorithm continues by identifying the $R$ leading eigenvectors (see phase \emph{subspace}), which indeed converge to $\b W$.
To support correlated features, we estimate the covariance matrix and apply pre/post-processing stages. Finally, we use decoupled standard linear ridge regressions to construct $\b V_i$.

% One sentence: Summery of the theoretical results
Appealingly, under reasonable statistical assumptions (see Section \ref{sec:theory} for details), the probability that our estimated $\bh W$ is far from the true $\b W$ can be bounded favorably as a function of the number of related tasks $I$, and the number of samples for each task $T$.

% One sentence: CMR is only initialziation, to fully solve use this optimization problem
In practice, $\bh W$ can be used as a stand-alone regressor or as an initialization for eq. (\ref{wv_opt_problem}).
% \begin{align} \begin{split} \label{wv_opt_problem}
% \min_{\b W, \b V_1, ... \b V_I}
%     &\sum_{i,t} \left( y_{it} - \tr{\b  W^T \b X_{it} \b V_i} \right)^2 .
% \end{split}\end{align}
Solving this non-convex optimization is non-trivial, but is much easier given an accurate initialization  \cite{candes2015phase, netrapalli2013phase,tu2015low,vaswani2017low}. It can be minimized using gradient descent or via alternating least squares techniques that solve for $\b W$ while fixing $\b V_i$ and vice versa. Either way, there is an inherent ambiguity in $\b W$ and $\b V_i$, as the outcome is invariant to transformation of the form $\b W \rightarrow \b W \b Z$ and $\b V_i \rightarrow \b V_i \b Z^{-T}$ for any invertible matrix $\b Z$. Thus, it is necessary to add an orthogonality constraint to $\b W$. Finally, a ridge regularization with respect to $\b V_i$ is needed, as we are typically interested in problems with a small value for $T$.% as in (\ref{ridge}). 

\section{Theoretical Analysis of CMR} 
\label{sec:theory}
We now theoretically analyze the intuitive CMR algorithm described in the previous section. 
In addition to the \emph{CMR model} assumption, we make the following statistical assumptions on the data:
\begin{enumerate}[label={[A\arabic*]}]

  \item \textbf{Matrix Variate Normal}: similarly to \cite{zhang2010learning,stegle2011efficient,bonilla2008multi}, we use the Kronecker model assumption for our multitask approach. We assume that $\b X_{it}$ are randomly chosen from a per-task zero-mean \textit{matrix variate normal distribution}:
  \begin{equation} \nonumber \begin{gathered}
      \mathbb E\left[ \b X_{it} \right] = 0 \quad ; \quad
      \text{cov}(\text{vec}(\b X_{it})) = \b \Gamma^{(i)} \otimes \b \Delta^{(i)}
  \end{gathered}\end{equation}
  for $\b \Gamma^{(i)} \in \mathbb R^{B \times B}$ and $\b \Delta^{(i)} \in \mathbb R ^ {P \times P}$, which will be referred to as $B$-covariance and $P$-covariance respectively. In matrix notations, the Kronecker structure implies that
  $\mathbb E\left[\b X_{it}\b X_{it}^T\right]= \b \Gamma^{(i)}. $
  \vspace{0.2cm}
  \item \textbf{Shared $B$-covariance}: we further assume the $B$-covariance will be shared across all the regression tasks, i.e. 
  $ \b \Gamma^{(i)} = \b \Gamma. $
\end{enumerate}

Given these natural assumptions, we are ready to state our main theoretical result which characterizes the quality of our spectral algorithm as a function of the relationship between the number of tasks $I$, the number of samples per task $T$, the feature dimensions $B$ and $P$, and the reduced rank $R$:

\begin{theorem} \label{correctness_thrm}
Under assumptions [A1]-[A2], if it holds that $R \leq T$ and in addition,
\begin{equation} \label{theorem_IT_req}
    IT \geq k \cdot \epsilon^{-2} \cdot R^4 \cdot 
    \max \left \{\frac {\log(IT)} T, 1 \right\}^2 \cdot
    \max\left\{\frac PT, 1\right\} \cdot B
\end{equation}
for some constant $k$ (that depends on the matrices $\b \Gamma, \b \Delta^{(i)}, \b W, \b V_i$), then we are guaranteed that
\begin{equation} \nonumber
\mathbb P \left[ \dist{\b W, \bh W} \geq \epsilon\right] 
    \leq \frac 1B + 9e^{-B}.
\end{equation}
\end{theorem}

% One sentence: An explanation on the theoretical results
%As one might expect, the theorem shows that $\b W$ can be efficiently recovered given $T \gg C(P + B/I)$ samples for a constant $C$ that depends on $R$. Furthermore, a large number of tasks ($I$) can compensate for a small number of samples per task ($T$). A surprising yet useful consequence is that in certain settings it is possible to recover $\b W$ when $T<P$, even though it is impossible to fully recover the $\b V_i$'s. In hindsight this is also intuitive: even weak signals that are not enough to fully characterize local behavior still inform the common mechanism.

% One sentence: An explanation on the theoretical results
The theorem above shows that a large number of tasks ($I$) can compensate for a small number of samples per task ($T$), which agrees with previous results in non-spatio-spectral MTL settings \cite{crammer2012learning, ben2003exploiting}. Furthermore, the theorem shows that $\b W$ can be efficiently recovered given $T \gg C(P + B/I)$ samples for a constant $C$ that depends on $R$. A surprising yet useful consequence is that in certain settings it is possible to recover $\b W$ when $T<P$, even though it is impossible to fully recover the $\b V_i$'s. In hindsight this is also intuitive: weak signals that are not enough to fully characterize local behavior can still inform the common mechanism.

It is also instructive to consider special cases of the theorem. When $I=1$, CMR reduces to a classic low-rank matrix recovery problem, and theorem \ref{correctness_thrm} suggests that the parameters can be recovered when $T \gg C(B+P)$ for some constant $C$. When $T=1$ and $P=1$, CMR reduces to a phase-retrieval variant, and the theorem suggests that recovery requires $ T\gg C \cdot B\log I$. Both results are consistent with previous analyses \cite{recht2010guaranteed, candes2015phase, netrapalli2013phase}.

The main idea of the proof of our results is as follows. Recall that the eigenspace of $\bh B$ characterizes the common component $\bh W$. It turns out that, with high probability, $\bh B$ is concentrated around its expectation. Ignoring correlations and constants, this expectation is proportional to $\b W \b W^T + \frac PT \b I$, thus $\b W$ coincides with its leading $R$-dimensional eigenspace. 
% we use Davis Kahan $\sin \Theta$ theorem \cite{yu2014useful} to quantify the distance between the eigenspaces as a function of the distance between $\bh A$ and its expectation.
It can be seen that the expectation of $\bh B$ is bounded from below (in the positive semi-definite sense) by the constant before the identity, which is inversely proportional to $T$. Indeed, this dependency is unique to our work and is a direct consequence of the double averaging with respect to both $T$ and $I$. This is also what allows eliminating the logarithmic factor found in the standard analysis of the spectral initialization \cite{netrapalli2013phase, candes2015phase}. See appendix A in the supplementary material for full details of the proof.

% One sentence: Why RIP is not relevant here.
%In the context of existing low rank minimization results, previous works suggest that optimization problems that satisfy certain restricted isometry properties (RIP) can be solved with standard descent algorithms and do not suffer from spurious local minima \cite{recht2010guaranteed, ge2017no, tu2015low}. When the number of samples is large enough and the features satisfy the $i.i.d$ assumption, it can be shown that these conditions are satisfied with high probability. It is non-trivial to extend these to the case of $I$ multiple tasks, with correlated block-structures as defined in (\ref{Mdef}). Furthermore, our synthetic numerical results indicate the existence of bad local minima, which can be avoided using CMR. %such analysis could not be applied in this case, as the matrices $\b M_{it}$ are not elementwise i.i.d due to the very nature of the shared $\b W$.

%%%%%%%%%%%%%%%%%%%%%%%%%%%%%%%%%%%%%%%%%%%%%%%%%%%%%%%%%%%%%%%%%%%%%%%%%%%%%%%%%%%%%%%%%%%%%%%%%%%%%%%%%%%%%%%%%%%%%%%%%%%%%%%%%%%%%%%%%%%%%%%
\section{Experimental Evaluation}
\label{sec:experiments}

We now demonstrate the efficacy of our approach on synthetic data, a simple convolution network image classification scenario, and the challenging real-life setting that motivated CMR, namely that of discharge estimation at multiple river locations. 

\subsection{Synthetic Data Simulations}
%%%%%%%%%%%%%%%%%%%%%%%%%%%%%%%%%%%%%%%

\begin{figure*} [h]
\centering
\begin{tabular}{cc}
\includegraphics[trim=12 0 100 20, clip, height=0.28\textwidth]{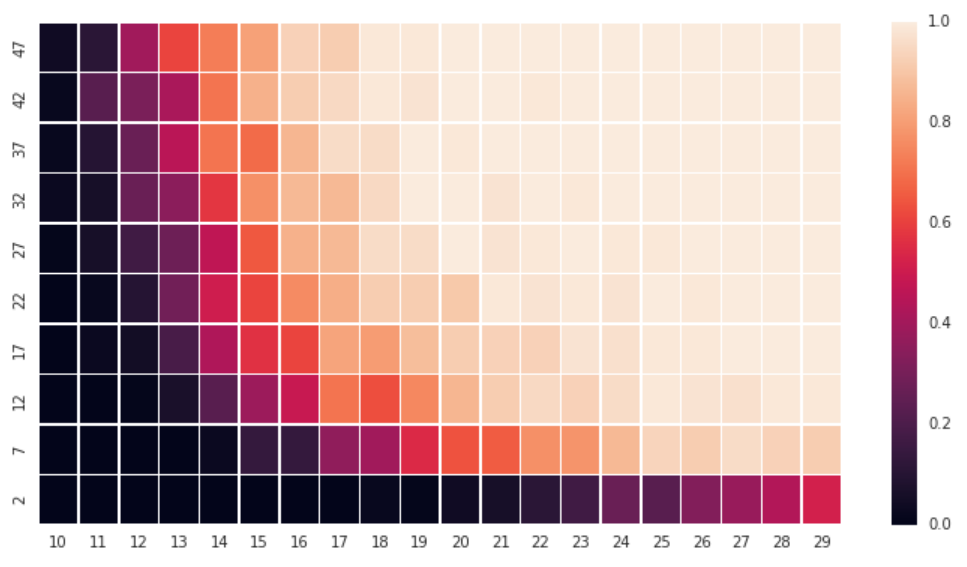} &
\includegraphics[trim=0 0 0 15, clip, height=0.28\textwidth]{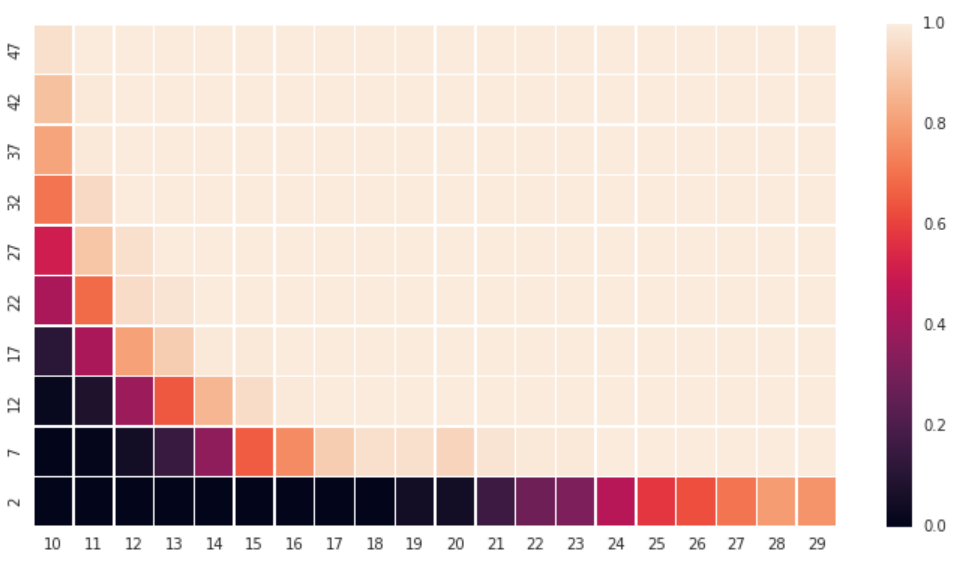}
\end{tabular}
%\vspace{-0.1in}
\caption{Observed probability for recovery of the true shared mechanism $\b W$ as a function of the number of tasks $I$ (y-axis) and the number of samples per task $T$ (x-axis) without (left) and with (right) CMR as initialization on synthetically generated data.}
\label{fig:CMR_synth}
\end{figure*}

We start by assessing the merit of CMR in a synthetic setting. Recall that our goal is to leverage measurements from many locations to improve prediction. Thus, we consider the performance of CMR for a range of values for the number of tasks $I$ and the number of samples per task $T$. For each set of $I, T$, we repeat the following 50 times: choose a random $\b W$ and $\b V_i$, solve (\ref{wv_opt_problem}) using gradient decent, and declare success if the squared correlation between the true $\b W$ and its estimate exceeds $0.90$. 
We do this with and without CMR as initialization. For simplicity, we use $R=1$ and use $i.i.d$ samples, i.e. $\b \Gamma = \b I, \b \Delta^{(i)}=\b I$. The results for $B=20$ and $P=10$ are summarized in Figure \ref{fig:CMR_synth}, where light tiles correspond to a high success rate. 

As expected, the results demonstrate that we can recover $\b W$ with few samples per task, when there are many such tasks. Interestingly, we also succeed in recovering $\b W$ when $T<P$, a setting where it is impossible to recover $\b V_i$. Comparison of the left panel (without initialization) to the right one (with initialization) illustrates the importance of the CMR initialization, which substantially widens the region of success. Further synthetic experiments that demonstrate the dependency between the CMR algorithm's probability of success and $I, T$ and $B$ are consistent with the theoretic results and are provided in the supplementary material.

\subsection{Real Life Settings}
%%%%%%%%%%%%%%%%%%%%%%%%%%%%%%%%%%%%%%%%%%%%%%%%%%%%%%%%

We now consider several real-life settings. For all datasets, we compare the CMR algorithm, as described in algorithm 1, to the following baselines\footnote{Unless otherwise stated, baselines were implemented via TensorFlow \cite{tensorflow2015-whitepaper}}:

\begin{itemize}%[leftmargin=0.8cm]
  \item {\bf{FRR}} - Full per-task ridge regression from the pixels in all bands/channels to the labels \cite{scikit-learn}.
%  \item {\bf{MTTE}} - Task embedding approach, where a cross-tasks ridge regression with task embedding features is performed. In this intuitive baseline, we first concatenate each sample features with a one-hot embedding of its specific task, and perform one ridge regression over all the samples of all the tasks together. To allow more complex non linear features, we follow \cite{huang2006extreme, rahimi2009weighted} and perform a random non linear transformation over the features. This approach is closely related to csMTL \cite{silver2008inductive, zhang2017survey}, where the main difference is that we use random non linear features instead of training a neural network.
  \item {\bf{TNR}} - Multitask learning via trace norm regularization \cite{pong2010trace, zhang2017survey}.
  \item {\bf{MTFS}} - Multitask feature selection approach, also referred as multitask lasso \cite{zhang2017survey, obozinski2006multi, liu2009multi}.
\end{itemize}
In addition, we compare the following CMR variants:
\begin{itemize}
  %\item {\bf{RND}} - Choosing a random $\b W$ (independently generated per fold), and performing Ridge regression to find $\b V_i$ as shown in stage \emph{regressions} of Algorithm 1. 
  \item {\bf{CMR-NW}} - The CMR model, when skipping the whitening phase, i.e. assuming $\b \Gamma = \b I$. By comparing this model to CMR, one can see the advantages of the CMR whitening phase.
  \item {\bf{CMR1}} - The CMR model when applied independently for each task so that $I=1$. The difference between this model and CMR illustrates the benefit that comes from transfer learning.
\end{itemize}

%As discussed, in principle additional descent steps on eq. (\ref{wv_opt_problem}) can be performed following the CMR algorithm output. However, the effect on our results was quite small and thus, for simplicity, we report the results of CMR as is.

\subsubsection{Multi-class Image Classification}
%%%%%%%%%%%%%%%%%%%%%%%%%%%%%%%%%%%%%%%%%%%%%%%%%%%%%%%%
We start by considering two simple multi-class image classification tasks \cite{simonyan2014very, krizhevsky2012imagenet}. For our purposes, we divide these into multiple binary classification tasks. In reality, the datasets below have enough samples for per-task learning, and thus we use them mostly to exemplify the power of our approach while being able to \emph{control} for the number of available samples. 

The motivation for using CMR in the image classification setting is that the standard architecture for such problems involves multiple neural layers that construct features, e.g., edge detectors, followed by a per-class linear front-end \cite{krizhevsky2012imagenet, simonyan2014very}. Similarly, CMR applies a common $\b W$ to identify shared features, followed by per-task linear regressions. 

To apply CMR on images, we use the process as illustrated in Figure \ref{cmr_classification_drawing}. We start by a convolution-like reshaping in which each image is divided to $P$ non-overlapping blocks containing $B$ pixels each, and is ordered as a tensor of $\sqrt{P}\times\sqrt{P}$ pixels with $B$ bands/channels (in practice, we just use $B \times P$ matrices). Since linear classifiers are insufficient for most tasks, we follow \cite{huang2006extreme, rahimi2009weighted} and up-lift the bands dimension using random projections and non-linear rectified linear units. These random features can also be replaced by other pre-trained convolution networks. Next, the shared $\b W$ transformation converts the $\sqrt{P}\times\sqrt{P} \times B$ matrix into an $\sqrt{P}\times\sqrt{P} \times R$ matrix. Finally, the per-class linear $\b V_i$ classifier is applied per task (with additional intercept and ridge parameters). 

{\bf{MNIST dataset}}:
%%%%%%%%%%%%%%%%%%%%%%%%%%%%%%%%%%%%%%%%%%%%%%%%%%%%%%%%
We start with a simple setting based on the MNIST \cite{lecun1998gradient} dataset, consisting of 70000 grey-scale images of dimension $28 \times 28$ of handwritten digits and their labels. To illustrate the effectiveness of CMR and control the number of individual classifications tasks, we consider a toy setting of 45 binary classification problems for every pair of digits. Each $28 \times 28$ image is reshaped to $7 \times 7$ images with $16$ bands, and uplifted in a non-linear manner to $B = 100$.

{\bf{SVHN dataset}}:
%%%%%%%%%%%%%%%%%%%%%%%%%%%%%%%%%%%%%%%%%%%%%%%%%%%%%%%%
We also consider the more challenging Street View House Numbers (SVHN) dataset \cite{netzer2011reading}. The data structure is similar where each $32\times 32 \times 3$ RGB image is reshaped to $8\times 8$ blocks of $48$ bands, and uplifted non-linearly to $B=200$.
%The results are summarized in Table $\ref{SVHN_result}$, and demonstrate a similar advantage to our CMR model. Similarly to the case of the MNIST dataset, experiments with CMR-NW led to significantly worse predictions.

Table \ref{MNIST_result} compares the performance of CMR with its competitors for both datasets. As expected, in both cases, the ability of CMR to perform transfer learning is evident, particularly when the number of training examples per task (T) is small. 
This is most apparent when CMR is compared to classic MTL algorithms that are not aware of the inherent two dimensionality of the features.
Experiments with CMR-NW led to inferior results and thus were omitted for brevity.

\begin{table}[h]
    %\vskip -0.4cm
    \caption{
        Comparison of the different algorithms on the MNIST and SVHN datasets. Shown is the average classification accuracy across $10$ random train/test repetitions. All the differences are statistically significant by more than two standard deviations.
    }
    \vskip 0.5cm
    \label{MNIST_result}
    \label{SVHN_result}
    \centering
    
    \footnotesize
    \begin{tabular}{cc}
    %\toprule
        \begin{tabular}{lccccc}
        \multicolumn{6}{c}{\textbf{MNIST}}                   \\
    
            \toprule
            & CMR & CMR1 & FRR & TNR & MTFS \\
            \midrule
                T=50      & $96.9$    & $95.5 $   & $95.6$   & $95.4$  & $94.7$ \\
                T=100     & $97.8 $   & $96.4 $   & $96.5 $  & $96.2$  & $95.8$ \\
                T=1000    & $98.92 $  & $98.13 $  & $98.34$  & $98.22$ & $98.36$ \\
            \bottomrule
        \end{tabular}
    &
        \begin{tabular}{lccccc}
            \\
            %\\
            \multicolumn{6}{c}{\textbf{SVHN}}                   \\
            \toprule
            & CMR & CMR1 & FRR & TNR & MTFS \\
            \midrule
                T=100      & $75.0 $    & $67.0 $  & $54.2$ & $59.8$ & $58.2$ \\
                T=200      & $81.1 $    & $75.6 $  & $58.6$ & $65.1$ & $63.2$\\
                T=400      & $85.1 $    & $81.3 $  & $65.2$ & $70.1$ & $68.2$\\
                T=800      & $87.7 $    & $84.5$   & $72.5$ & $73.7$ & $72.0$\\
            \bottomrule
        \end{tabular}
    \end{tabular}
\end{table}

\subsubsection{River Discharge Estimation}
%%%%%%%%%%%%%%%%%%%%%%%%%%%%%%%%%%%%%%%%%%%%%%%%%%%%%%%%

Finally, we assess the benefit of our CMR approach for the motivating task of remote discharge estimation. We use images from the LANDSAT8 mission \cite{roy2014landsat} that include 11 spectral bands each, and ground truth labels from the United States Geological Survey (USGS) website. We consider the 1000 sites across the U.S. that have the most samples and discard sites for which the prediction problem is trivial, i.e. where month-average predictor has less than 10\% error. The resulting mean squared errors (MSE) over 40 randomly shuffled folds are reported in Table \ref{discharge_result}. As before, the advantage of CMR over CMR1, FRR, TNR and MTFS is evident, especially when $T$ is small, as is common in remote sensing. That said, we note that the results are less significant due to the noisy and unbalanced nature of the dataset, with few extreme-value events.
\begin{table}[ht]
\caption{Comparison of the different methods for the real-life task of remote discharge estimation in multiple river locations. Shown is the average MSE of the log-discharge, normalized per task. The results below are less significant than the results in the other datasets, where the differences are at the order of 1.3 standard deviations.}

\label{discharge_result}
\vspace{0.5cm}
\centering
\begin{tabular}{lccccc}
    \toprule
    & CMR & CMR1 & FRR & TNR & MTFS \\
    \midrule
        T=40      & $40 $ & $46$    & $52$       & $55$ &  $51$ \\
        T=60      & $29 $ & $31$    &  $34$      & $35$ &  $33$ \\
        T=80      & $27 $ & $30$    &  $31$      & $32$ &  $31$ \\
        T=100     & $26 $ & $29$    &  $29$      & $29$ &  $29$ \\
    \bottomrule
\end{tabular}
\end{table}

\begin{figure*}[t]
\begin{center}
\hspace{-0.4cm}
\begin{tabular}{l}
\includegraphics[height=0.175\textwidth]{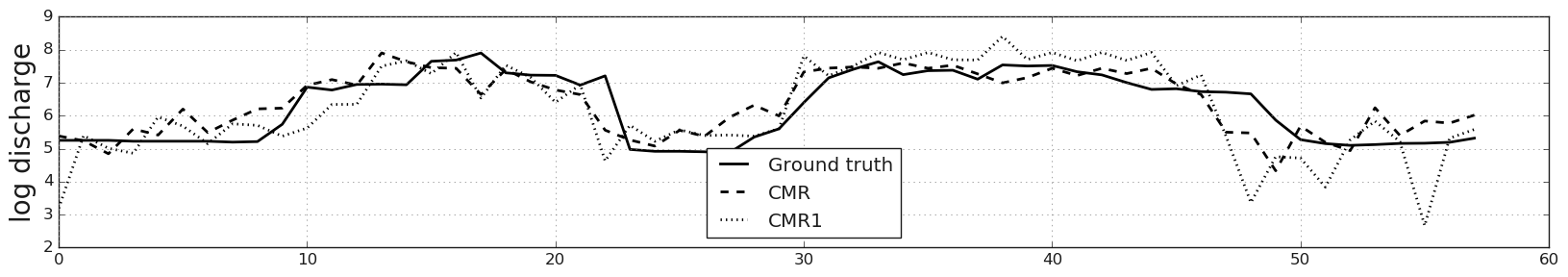} \\
\vspace{-0.2cm} \hspace{-0.2cm}
\includegraphics[height=0.195\textwidth]{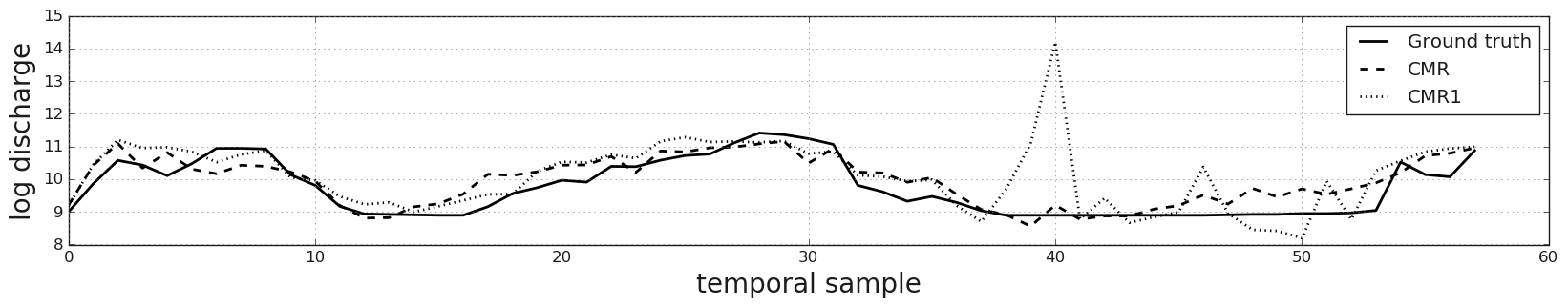}
\end{tabular}
\end{center}
\vspace{-0.3cm}
\caption{Examples of the true log-discharge (solid black) as well as prediction by our CMR model (dashed black) and the baseline CMR1 model (dotted line) for two different river sites}
%\vskip -0.2cm
\label{example_discharge}
\end{figure*}

To gain a qualitative sense of the nature of the results, Figure \ref{example_discharge} shows the true and predicted log-discharge for several river sites. For readability, we only present the true discharge, and the predictions by our CMR models as well as the CMR1 baseline. As can be expected, CMR1, which does not benefit from transfer learning is always noisier than CMR. This leads to inferior performance on average (as is evident in the top two panels) or good results in some regions but substantial failures in others as is exemplified in the bottom panel.

Finally, recall that the motivation for the common mechanism $\b W$ was the shared physical characteristics of spectral water reflection. To get a sense of what was actually learned, Figure \ref{patna_fig} shows the application of the estimated $\b W$ to an image in the Patna region. It is quite clear that our approach was able to automatically learn an effective "water detector".

%%%%%%%%%%%%%%%%%%%%%%%%%%%%%%%%%%%%%%%%%%%%%%%%%%%%%%%%%%%%%%%%%%%%%%%%%%%%%%%%%%%%%%%%%%%%%%%%%%%%%%%%%%%%%%%
\section{Summary and Future Work}

In this work, we tackled the challenge of leveraging few data points from multiple related regression tasks, in order to improve predictive performance across \emph{all} tasks. We proposed a common mechanism regression model and a corresponding spectral optimization algorithm for doing so. We proved that, despite the non-convex nature of the learning objective, it is possible to reconstruct the common mechanism, even when there are not enough samples to estimate the per-task component of the model. In particular, we characterized a favorable dependence on the number of related tasks and the number of samples for each task. We also demonstrated the efficacy of the approach on simple visual recognition scenarios using random convolution-like and nonlinear features, as well as a more challenging remote river discharge estimation task. 

On the modeling front, it would be useful to generalize our CMR approach so as to also allow for robust and task-normalized loss functions. In terms of the theoretical analysis, it would be interesting to also consider the conditions for satisfying RIP in the CMR model. 

\section{Acknowledgements}

This work was supported in part by the Israel Science Foundation (ISF) under Grant 1339/15.

\bibliographystyle{plain}
\bibliography{references}

\clearpage
\section*{Appendix A - Main theorem proof}

To simplify the notation, we start by defining the following:
\begin{gather} \label{L_D_M_L_def}
L_i \equiv \Ln \b \Gamma \Rn_2^2 \Ln \b W \Rn_2^2  \Ln \b \Delta_i \Rn_2^2 \Ln \b V_i \Rn_2^2, \\
D \equiv \frac 1I \sum_i L_i^2, \quad M \equiv \sqrt{\frac 1I \sum_i L_i^4},\quad
L \equiv \max_i L_i
\end{gather}

The proof of theorem 1 relies on several lemmas that are stated precisely below. The first shows that with high probability, $\bh A$ which defined in algorithm 1 to be:
$$ \bh A \equiv \frac 1{IT^2}\sum_{itt'} y_{it} y_{it'} \b X_{it} \b X_{it'}^T $$
is, with high probability, close to its expectation.

%%%%%%%%%%%%%%%%%%%%%%%%%%%%%%%%%%
%% A Concentration Lemma
%%%%%%%%%%%%%%%%%%%%%%%%%%%%%%%%%%

\begin{lemma} \label{A_expectation}
The expectation of $\bh A$ satisfies:
\begin{align}
\mathbb E\left[ \bh A \right] 
    = \left(1 + \frac 1T \right) \b \Gamma  \b W \b Q \b W^T \b \Gamma  + \beta \cdot \b \Gamma
    \equiv \b A
\end{align}
where $\b Q \in \mathbb R^{R \times R}$ and $\beta \in \mathbb R$ are defined by
\begin{gather}
\b Q = \frac 1I \sum_i \b V_i^T \b \Delta_i^2 \b V_i
, \qquad
\beta = \frac 1T \tr{\b W^T \b \Gamma \b W
                                \left( \frac { \sum_i \b V_i^T \b \Delta_i \b V_i \tr {\b \Delta_i}} I \right) },
\end{gather}
and in addition, if it holds that 
\begin{equation} \label{A_expecation_C_def}
    IT \geq C(\epsilon) \cdot R^3  \cdot
                     \max\left\{\frac RT, 1\right\}^2 \cdot 
                     \max \left \{\frac {\log(IT)} T, 1 \right\}^2 \cdot
                     \max\left\{\frac PT, 1\right\} \cdot
                     B 
\end{equation}
for
\begin{equation}
    C(\epsilon) \equiv c_1 D \epsilon^{-2} + c_2 M \epsilon^{-2} + c_3 \max \left\{L \epsilon^{-1}, L^2 \epsilon^{-2} \right\},
\end{equation}
$c_1, c_2, c_3$ are some absolute constant and $D, M, L$ are defined in eq. (\ref{L_D_M_L_def}), then it holds that:
\begin{equation}
\P{\Ln \b A - \bh A \Rn_2 \geq \epsilon} \leq 5 e^{-B} + \frac 1B
\end{equation}

\end{lemma}

The second Lemma shows that the estimation of the $B$-covariance $\bh \Gamma$, which is defined in algorithm 1 to be:
$$
\bh \Gamma \equiv \frac 1{ITP} \sum_{it} \b X_{it} \b X_{it}^T
$$
is, with high probability, close to its expectation $\b \Gamma$.
\vspace{0.05in}

%%%%%%%%%%%%%%%%%%%%%%%%%%%%%%%%%%
%% Cov Concentration Lemma
%%%%%%%%%%%%%%%%%%%%%%%%%%%%%%%%%%

\begin{lemma} \label{lambda_close_lemma}
If it holds that:
\begin{equation}
    ITP \geq D(\epsilon) B
\end{equation}
where
\begin{gather} \label{lambda_close_lemma_D_def}
    D(\epsilon) \equiv D'\max\left\{ \kappa_{\b \Gamma} \bar K \epsilon^{-2}, 
                                     \kappa_{\b \Gamma}^2 K_{max}\epsilon^{-1} \right\}, \\
\label{K_bar_max_def} \bar K \equiv \frac 1I \sum_i \Ln \b \Delta_i \Rn_2 ^2
, \quad 
K_{max} \equiv \max_i \Ln \b \Delta_i \Rn_2,
\end{gather}
$D'$ is some absolute constant and $\kappa_{\b \Gamma}$ is the condition number of $\b \Gamma$ , then it holds that:
\begin{equation}
\P{\Ln \b\Gamma - \hat{\b \Gamma} \Rn_2 \geq \lambda_{\min} (\b \Gamma) \epsilon}
        \leq e^{-B}
\end{equation}
\end{lemma}

Using the two above lemmas, it can be seen that if both $\bh A$ and $\bh \Gamma$ are equal to their expectation, we get that:
$$
\bh B \equiv \bh \Gamma^{-1/2} \bh A \bh \Gamma^{-1/2} 
    = \left(1 + \frac 1T \right) \b \Gamma^{1/2} \b W \b Q \b W^T \b \Gamma^{1/2} + \beta \b I
$$
and thus
$$
\text{span}\left[ \bh W \right] 
    \equiv \text{span} \left[ \bh \Gamma^{-1/2} \text{eigv}_R\left( \bh B \right)\right] 
    = \text{span}[\b W].
$$
Hence, in the third lemma we use the Davis-Kahan $\sin\Theta$ theorem \cite{yu2014useful} to quantify how much a deviation of $\bh A$ and $\bh \Gamma$ from their expectation affects the deviation of $\bh W$ from $\b W$:

\begin{lemma} \label{khan_lemma}
If for some $\epsilon_1$ and $\epsilon_2$ it holds that
\begin{equation} \nonumber
\Ln \bh \Gamma - \b \Gamma \Rn_2 \leq\lambda_{\min}(\b \Gamma) \epsilon_1 \quad \text{and} \quad \Ln \bh A - \b A \Rn_2 \leq \epsilon_2
\end{equation}
then it holds that
\begin{equation} \nonumber
\dist{\bh W, \b W} \leq f(\epsilon_1, \epsilon_2) 
\end{equation}
where $\kappa_{\b \Gamma}$ is the condition number of $\ \b \Gamma$, and
\begin{align} \nonumber \begin{split}
    f(\epsilon_1, \epsilon_2) 
        \equiv \frac {\epsilon_1} {1 - \epsilon_1}
                + 2 \sqrt R \left( \frac {\kappa_{\b \Gamma}+ \epsilon_1} 
                                     {1 - \epsilon_1} \right)^{1.5} 
        \frac {\epsilon_2 + \beta \lambda_{\min}(\b \Gamma) \epsilon_1}
                    {\lambda_{\min}\left(\b \Gamma \b W \b Q \b W^T \b \Gamma \right)}
\end{split}\end{align}
\end{lemma}
Using these three lemmas, we can now safely continue the proof. We start by defining $\epsilon_1$ and $\epsilon_2$ similarly to their definition in lemma \ref{khan_lemma}:
\begin{equation}
    \Ln \bh \Gamma - \b \Gamma \Rn_2 \equiv \lambda_{\min}(\b \Gamma) \epsilon_1 \quad ; \quad \Ln \bh A - \b A \Rn_2 \equiv \epsilon_2
\end{equation}
For some $\gamma \geq 0$, we use lemma \ref{A_expectation} and lemma \ref{lambda_close_lemma} to see that if it holds that
\begin{equation} \label{theorem_IT_cond_with_k_tag}
    IT \geq k' \cdot \gamma^{-2} \cdot R^4 \cdot 
    \max\left\{\frac RT, 1\right\}^2 \cdot 
    \max \left \{\frac {\log(IT)} T, 1 \right\}^2 \cdot
    \max\left\{\frac PT, 1\right\} \cdot B
\end{equation}
where $k'$ is a constant that depends on the matrices $\b \Gamma, \b \Delta^{(i)}, \b W, \b V_i$ (see section \ref{k_expansion_section} for more details), then we are guaranteed that the following conditions
\begin{enumerate}
    \item $\epsilon_1 \leq \frac 12 $
    \item $\epsilon_1 \leq  \frac T{\eta R\sqrt R P} \cdot \gamma$
    \item $\epsilon_2 \leq \frac {\lambda_{\min}\left(\b \Gamma \b W \b Q \b W^T \b \Gamma \right) } {\sqrt R} \cdot \gamma$
\end{enumerate}
hold with probability of:
\begin{equation} \label{prob_of_conditions}
\P{(1) \wedge (2) \wedge (3) } \geq 1 - \frac 1B - 9 e^{-B} .
\end{equation}
Under the event that (1), (2) and (3) hold, we can use lemma \ref{khan_lemma} to bound the distance in the following way:
\begin{align} \label{dist_result_with_gamma}
    \dist{\bh W, \b W} 
    &\leq \frac {\epsilon_1} {1 - \epsilon_1}
                + 2 \sqrt R \left( \frac {\kappa_{\b \Gamma}+ \epsilon_1} 
                                     {1 - \epsilon_1} \right)^{1.5} 
        \frac {\epsilon_2 + \beta \lambda_{\min}(\b \Gamma) \epsilon_1}
                    {\lambda_{\min}\left(\b \Gamma \b W \b Q \b W^T \b \Gamma \right)} \leq c \cdot \gamma
\end{align}
where $c$ is a constant that depends solely on the condition number of $\b \Gamma$
and satisfies $c > 1$. 
By defining $\epsilon \equiv c\cdot \gamma$ and $k \equiv c^2 \cdot k'$, we get from eq. (\ref{theorem_IT_cond_with_k_tag}), (\ref{prob_of_conditions}) and (\ref{dist_result_with_gamma}) that if $R < T$ and
\begin{equation}\label{theorem_IT_cond}
    IT \geq k \cdot \epsilon^{-2} \cdot R^4 \cdot 
    \max \left \{\frac {\log(IT)} T, 1 \right\}^2 \cdot
    \max\left\{\frac PT, 1\right\} \cdot B
\end{equation}
then it holds that
\begin{equation}
    \P{\dist{\bh W, \b W} \geq \epsilon } \leq \frac 1B + 9e^{-B}.
\end{equation}
\qed

Below are the detailed proofs for lemma \ref{A_expectation} and lemma \ref{lambda_close_lemma}
, followed by the expansion of the constant $k$ from eq. (\ref{theorem_IT_cond})
. For reasons of brevity we decided to omit the proof of lemma \ref{khan_lemma} since it mostly consists of algebraic manipulations.

%%%%%%%%%%%%%%%%%%%%%%%%%%%%%%%%%%%%%%%%%%%%%%%%%%%%%%%%%%%%%%%%%%%%%%%%%%%%%%%5
%% Proof of A Concentration
%%%%%%%%%%%%%%%%%%%%%%%%%%%%%%%%%%%%%%%%%%%%%%%%%

\subsection{Proof of Lemma \ref{A_expectation}}

The proof below is split to several major parts:
\begin{itemize}
    \item We first change $\bh A$ to be defined with standard Gaussian matrices $\b K_{it}$ with independent entries as opposed to the correlated $\b X_{it}$.
    \item We split $\bh A$ to three different elements, and bound the distance of each one from its expectation.
    \item We use the union bound to show that $\bh A$ is close to its expectation.
\end{itemize}

%%%%%%%%%%%%%%%%%%%%%%%%%%%%%%%%%%%%%%%%%%%%%%%%%%%%%%%
\subsubsection{Standardize}
%%%%%%%%%%%%%%%%%%%%%%%%%%%%%%%%%%%%%%%%%%%%%%%%%%%%%%%
First, according to the assumptions, $\b X_{it} \sim \mathcal N(\b 0_{B\times P}, \b \Gamma \otimes \b  \Delta_i)$ and thus can be rewritten as follows:
\begin{equation}
\b X_{it} = \b \Gamma^{1/2} \b R_{it} \b \Delta_i^{1/2} 
    \quad ; \quad 
\b R_{it} \sim \mathcal {N}(\b 0_{B \times P},\b I_B \otimes \b I_P)
\end{equation}
where $\b R_{it}$ are random matrices with independent standard Gaussian elements. Using that we can write:
\begin{equation}
\tr{\b W^T \b X_{it} \b V_{i}} 
    = \tr{\left(\b \Gamma^{1/2}\b W\right)^T \b R_{it} \left( \b \Delta_i^{1/2}\b V_{i}\right) }
\end{equation}
and using SVD decomposition, the following can be denoted:
\begin{align}
\b \Gamma^{1/2} \b W = \b B_{W} \b \Sigma_{W} \b C_W
    \quad &\text{where} \quad \b B_W  \in \mathbb R^{B\times B}, 
                                                 \b \Sigma_W \in \text{Diag}^{B \times R},
                                                 \b C_W \in \mathbb R^{R\times R} \\
\b \Delta_i^{1/2} \b V_i = \b B_{V_i} \b \Sigma_{V_i} \b C_{V_i} 
    \quad &\text{where} \quad \b B_{V_i}  \in \mathbb R^{P\times P}, 
                                                 \b \Sigma_{V_i} \in \text{Diag}^{P \times R},
                                                 \b C_{V_i} \in \mathbb R^{R\times R}
\end{align}
where $\b B_{W}, \b B_{V_i}, \b C_{W}, \b C_{V_i}$ are orthogonal matrices.  It can be seen that $\b R_{it}$ is a random matrix with independent Gaussian elements thus is invariant to multiplication by orthogonal matrices both from the left and from the right. We can use this rotation invariance and the fact that  $\b B_{W}$ and $\b B_{V_i}$ are constants and are chosen before $\b R_{it}$ is drawn, to assume that the matrices $\b R_{it}$ are chosen in the following way:
\begin{equation}
\b R_{it} \equiv \b B_W^{T} \b K_{it} \b B_{V_i}^{T} \quad ; \quad \b K_{it} \sim \mathcal N (\b 0_{B\times P}, \b I_B \otimes \b I_P) 
\end{equation}
and when putting it together,
\begin{align}
\b X_{it} &\equiv \b \Gamma^{1/2} \b B^T_W \b K_{it} \b B_{V_i}^T \b \Delta_i^{1/2} \\
\tr{\b W^T \b X \b V_i} 
        &= \tr{(\b \Sigma_{W} \b C_W)^T \b K_{it} (\b \Sigma_{V_i} \b C_{V_i}) } 
        \equiv \tr{\bb W^T \b K_{it} \bb V_i } \\
\b X_{it} \b X_{it'}^T
    &= \b \Gamma^{1/2} \b B^T_W \b K_{it} \b \Psi_i \b K_{it} \b B_W \b \Gamma^{1/2}
\end{align}
where $ \b \Psi_i \equiv \b B_{V_i}^T \b \Delta_i \b B_{V_i}, \ 
            \bb W \equiv \b \Sigma_W \b C_W$
and $\bb V_i \equiv \b \Sigma_{V_i} \b C_{V_i}$. When looking at $\bb W$ and $\bb V_i$, it can be seen that they are both block matrices with only the upper $R \times R$ block non-zero, thus can be denoted as:
\begin{equation} \label{WV_1_def}
\bb W \equiv \left( \begin{matrix} 
        \b W^{(1)} \in \mathbb R^{R \times R}  \\
        \b 0 \in \mathbb R^{(B-R) \times R}
    \end{matrix}\right) 
\quad ; \quad
\bb V_i \equiv \left( \begin{matrix} 
        \b V_i^{(1)} \in \mathbb R^{R \times R}  \\
        \b 0 \in \mathbb R^{(P-R) \times R}
    \end{matrix}\right) 
\end{equation}
and totally,
\begin{equation}
\Ln \b A - \bh A\Rn_2 \label{standart_A}
    \leq \Ln \b \Gamma \Rn_2 \Ln \frac 1 {IT^2} \sum_{itt'} y_{it} y_{it}' \b K_{it} \b \Psi_i \b K_{it}^T 
        - \left(1 + \frac 1T \right) \bb W \b Q \bb W^T  - \beta \b I \Rn_2
\end{equation}

%%%%%%%%%%%%%%%%%%%%%%%%%%%%%%%%%%%%%%%%%%%%%%%%%%%%%%%
\subsubsection{Split \texorpdfstring{$\bh A$}{Lg} to \texorpdfstring{$\b E_1, \b E_2$}{Lg} and \texorpdfstring{$\b E_3$}{Lg}}
%%%%%%%%%%%%%%%%%%%%%%%%%%%%%%%%%%%%%%%%%%%%%%%%%%%%%%%
We first define the following:
\begin{equation}
\b K_{it} = \left( \begin{matrix} 
        \b K_{it}^{(1)} \in \mathbb R^{R \times R}  & \b K^{(2)}_{it} \in \mathbb R^{R\times (P-R)}\\
        \b K^{(3)}_{it} \in \mathbb R^{(B-R)\times R} & \b K^{(4)}_{it} \in \mathbb R^{(B-R) \times (P-R)}
    \end{matrix}\right)
\end{equation}
and using this, we further define:
\begin{align}
&\bb K_{it} \equiv \left( \begin{matrix} 
        \b K_{it}^{(1)} \in \mathbb R^{R \times R}  & \b 0 \in \mathbb R^{R\times (P-R)}\\
        \b 0 \in \mathbb R^{(B-R)\times R} & \b 0 \in \mathbb R^{(B-R) \times (P-R)}
    \end{matrix}\right)  \\
&\bt K_{it} \equiv \left( \begin{matrix} 
        \b 0 \in \mathbb R^{R \times R}  & \b K^{(2)}_{it} \in \mathbb R^{R\times (P-R)}\\
        \b K^{(3)}_{it} \in \mathbb R^{(B-R)\times R} & \b K^{(4)}_{it} \in \mathbb R^{(B-R) \times (P-R)}
    \end{matrix}\right)
\end{align}
which totals to
\begin{equation}
\b K_{it} = \bt K_{it} + \bb K_{it}
\end{equation}
and we notice that
\begin{equation}
y_{it} = \tr{\bb W^T \b K_{it} \bb V_i} = \tr{\bb W^T \bb K_{it} \bb V_i} 
    = \tr{{\b W^{(1)}}^T \b K_{it}^{(1)} \b V_i^{(1)}}. 
\end{equation} 
We use this to get:
\begin{align}  \begin{split}
\frac 1{IT^2}\sum_{itt'} y_{it} y_{it'} \b K_{it} \b \Psi_i \b K_{it'}^T
    &= \frac 1{IT^2}\sum_{itt'} y_{it} y_{it'} (\bb K_{it} + \bt K_{it}) \b \Psi_i (\bb K_{it'} + \bt K_{it'})^T \\
    &= \frac 1{IT^2}\sum_{itt'} y_{it} y_{it'} \bb K_{it} \b \Psi_i \bb K_{it'}^T \\
           & \quad + \frac 1{IT^2}\sum_{itt'} y_{it} y_{it'} \bb K_{it} \b \Psi_i \bt K_{it'}^T
                    + \frac 1{IT^2}\sum_{itt'} y_{it} y_{it'} \bt K_{it} \b \Psi_i \bb K_{it'}^T \\
           & \quad + \frac 1{IT^2}\sum_{itt'} y_{it} y_{it'} \bt K_{it} \b \Psi_i \bt K_{it'}^T  \\
    &\equiv \b E_1 + \frac 12 \b E_2 + \frac 12 \b E_2 ^T + \b E_3
\end{split}\end{align}
and a direct consequence is:
\begin{align} \begin{split} \label{split_A} 
\Ln \frac 1 {IT^2} 
                \sum_{itt'} y_{it} y_{it'} \b K_{it} \b \Psi_i \b K_{it}^T 
        - \left(1 + \frac 1T \right) \bb W \b Q \bb W^T  - \beta \b I \Rn_2
    &\leq \Ln \b E_1 - \E{\b E_1} \Rn_2
        + \Ln \b E_2 - \E{\b E_2} \Rn_2 \\
        &\qquad + \Ln \b E_3 - \E{\b E_3} \Rn_2
\end{split}\end{align}

%%%%%%%%%%%%%%%%%%%%%%%%%%%%%%%%%%%%%%%%%%%%%%%%%%%%%%%
\subsubsection{Bound \texorpdfstring{$\b E_1$}{Lg}}
%%%%%%%%%%%%%%%%%%%%%%%%%%%%%%%%%%%%%%%%%%%%%%%%%%%%%%%

\begin{lemma}
If it holds that:
\begin{equation} \label{E_1_cond}
    IT \geq C_1(\epsilon) R^3 \max \left\{ \frac RT, 1\right\}^2 B
\end{equation}
where
\begin{equation} \label{C_1_def}
C_1(\epsilon) \equiv C_1' \cdot \frac {D} {\Ln \b \Gamma \Rn_2^2\epsilon^2} ,
\end{equation}
$C_1'$ is an absolute constant and $D$ is defined in eq. (\ref{L_D_M_L_def}),
% \begin{equation} \label{D_Li_definition}
%     D = \frac 1I \sum_i L_i^2 
% \quad ; \quad
%     L_i \equiv \Ln \b \Delta_i \Rn_2^2 \Ln \b V_i \Rn_2^2 \Ln \b \Gamma \Rn_2 \Ln \b W \Rn_2^2 
% \end{equation}
it holds that:
\begin{equation}
\P{\Ln \b E_1 - \E{\b E_1}\Rn_2 \geq \epsilon} \leq \frac 1 {3B}
\end{equation}
\end{lemma}

\begin{proof}

It can be noticed that $\b E_1$ involves 4'th Gaussian moment of full rank $R \times R$ random matrices. Since such matrices does not obey the sub-gaussian or sub-exponential laws, it is not straightforward to use Hoeffding/Bernstein inequalities to bound its deviation from its expectation. Thus, we use the fact that
\begin{equation}
    \Ln \b E_1 - \E{\b E_1} \Rn_2 \leq \Ln \b E_1 - \E{\b E_1} \Rn_F
\end{equation}
and the fact that Frobenious norm, as opposed $\ell_2$, has closed form, to calculate the variance of $\b E_1$ and bound its deviation using Chebysheff inequality. Using Isserli's theorem \cite{isserlis1918formula} and some straightforward (though rather tedious) algebra, it can be seen that the following bound holds:
\begin{equation}
    \E {\Ln \b E_1 - \E{\b E_1} \Rn_F^2} \leq \frac {c D R^3} 
        {\Ln \b \Gamma \Rn_2^2 IT} \cdot \max \left\{\frac R T, 1 \right\}^2
\end{equation}
where $c$ is an absolute constant and $D$ is defined in eq. (\ref{L_D_M_L_def}).
Thus, a direct consequence of Chebysheff inequality is
\begin{equation}
\P{\Ln \b E_1 - \E{\b E_1}\Rn_F \geq \epsilon} 
    \leq \frac {c D R^3} {IT \Ln \b \Gamma \Rn_2^2 \epsilon^2} 
        \cdot \max \left\{\frac R T, 1 \right\}^2
\end{equation}
In addition, if we define
\begin{equation}
C_1(\epsilon) \equiv  3 c \frac {D} {\Ln \b \Gamma \Rn_2^2 \epsilon^2}
\end{equation}
and we require
\begin{equation}
IT \geq C_1(\epsilon) R^3 \max \left\{ \frac RT, 1\right\}^2 B
\end{equation}
then it holds that
\begin{equation}
\P{\Ln \b E_1 - \E{\b E_1}\Rn_2 \geq \epsilon} \leq \frac 1{3B}
\end{equation}
\end{proof}

%%%%%%%%%%%%%%%%%%%%%%%%%%%%%%%%%%%%%%%%%%%%%%%%%%%%%%%
\subsubsection{Bound \texorpdfstring{$\b E_2$}{Lg}}
%%%%%%%%%%%%%%%%%%%%%%%%%%%%%%%%%%%%%%%%%%%%%%%%%%%%%%%
\begin{lemma}
If it holds that
\begin{equation} \label{E_2_cond}
IT \geq C_2(\epsilon) R^2 \max \left\{ 1, \frac R {\sqrt T} \right\} B
\end{equation}
where
\begin{equation} \label{C_2_def}% \label{M_def}
C_2(\epsilon) \equiv C_2' \cdot \frac {D+M}{\Ln \b \Gamma \Rn_2^2 \epsilon^2 } 
\end{equation}
$C_2'$ is an absolute constant and $D, M$ are defined in eq. (\ref{L_D_M_L_def}), then
\begin{equation}
\P{\Ln \b E_2 - \E{\b E_2}\Rn_2 \geq \epsilon} \leq \frac 1 {IT} +  e \cdot e^{-B}.
\end{equation}
\end{lemma}

\begin{proof}
To bound $\b E_2$, we first notice that Frobenious-based bounds are not tight enough and yield a result that is quadratic in $B$. Thus, we first assume that $\bb K_{it}$ (and thus $y_{it}$) are constant and bound $\b E_2$ given them, and then bound $\bb K_{it}$ separately. We can do it since $\bb K_{it}$ and $\bt K_{it}$ are independent. 

We start by noticing that
\begin{equation}
    \E{\b E_2} = 0 
\end{equation}
and thus, using the definition of the $\ell_2$ norm it holds that
\begin{equation}
\Ln \b E_2 - \E{\b E_2} \Rn_2  = \Ln \b E_2 \Rn_2  
    = \max_{\b u \in \mathbb B^B_{1}} \ \b u^T \b E_2 \b u
\end{equation}
where $\mathbb B^n$ is the unit-ball of $\mathbb R^n$. We follow the proof technique used in A.4.2 in \cite{candes2015phase} and in theorem 5.39 in \cite{vershynin2010introduction} and assume that $\b u$ is independent of $\b E_2$, and in the end we use a $\epsilon$-Net argument to prove it for $\b u$ that depends on $\b E_2$. It can be seen that
\begin{align}
\b u^T \b E_2 \b u
    &= \frac 2 {IT^2} \sum_{itt'} y_{it'} y_{it} 
                \left( \b u ^T \bb K_{it'} \b \Psi_i \bt K_{it}^T  \b u \right) \\
    &= \frac 2 {IT} \sum_{itbp}
                \left( \frac 1T \sum_{t'}^T \sum_{rr'}^R y_{it'} y_{it}  u_{r}
                                    K_{it'rr'} \Psi_{ir' p}  u_b \right)
                    \widetilde  K_{itbp} \\
    &\equiv \frac 2 {IT} \sum_{itbp} d_{itbp} \widetilde K_{itbp}
\end{align}
where $d_{itbp}$ depends solely on $\bb K_{it}$, and thus is independent of $\widetilde K_{itbp}$. Since $K_{itbp}$ are Gaussian, we can use Hoeffding inequality \cite{vershynin2010introduction} to bound its deviation
\begin{equation}  \label{E_2_given_u}
\Pgiven{\b u^T \b E_2 \b u \geq \epsilon} {\bb K_{it} }
    \leq e \cdot \exp\left[- c\frac { I T\epsilon^2} {\frac 1{IT} \sum_{itbp} d_{itbp}^2} \right]
\end{equation}
To give a bound for the extremal $\b u$, we use an $1/4$-net in the unit ball $\mathbb B^B$. According to lemma 5.2 in \cite{vershynin2010introduction}, to cover the unit ball in this vector space, such net needs a total of
\begin{equation} \label{1_4_net}
     | \mathcal N_{1/4} | = 9^{B}
\end{equation}
points, and using lemma 5.4 in \cite{vershynin2010introduction} we get that:
\begin{equation}
\max_{\b u  \in \mathbb B^{B}}\  \b u^T\b E_2  \b u
    \leq 2\max_{\b u \in \mathcal N_{1/4}}\ \b u^T \b E_2 \b u
\end{equation}
Since eq. (\ref{E_2_given_u}) holds for every constant $\b u$, we can use the union bound
\begin{align}
\Pgiven{2\max_{\bt y \in \mathcal N_{1/4}}\ \b u^T \b E_2 \b u \leq \epsilon} {\bb K_{it}}
    &\geq 1- e \cdot 9^B \cdot \exp\left[- c\frac { I T\epsilon^2} 
                                {\frac 1{IT} \sum_{itbp} d_{itbp}^2} \right] \\
    &\geq 1- e \cdot \exp\left[3B - c\frac { I T\epsilon^2} 
                                {\frac 1{IT} \sum_{itbp} d_{itbp}^2} \right]
\end{align}
and thus, if we require that:
\begin{equation}  \label{IT_given_d}
3B - c\frac { I T\epsilon^2} {\frac 1{IT} \sum_{itbp} d_{itbp}^2} \leq - B
    \quad \Longrightarrow \quad
IT \geq \frac 4c \left( \frac 1{IT} \sum_{itbp} d_{itbp}^2 \right) \epsilon^{-2} B
\end{equation}
we get that 
\begin{equation}
\Pgiven {\Ln \b E_2 \Rn_2 \geq \epsilon} {\bb K_{itbp}} \leq e \cdot e^{-B}
\end{equation}
\paragraph{Bounding \texorpdfstring{$\frac 1{IT} \sum_{itbp} d_{itbp}^2$:}{LG}}
To finish this bound, we need to show that with high probability, $\frac 1{IT} \sum_{itbp} d_{itbp}^2$ is bounded. to do so, we notice that
\begin{align}
\frac 1{IT}\sum_{itbp} d_{itbp} ^2
    &= \frac 1{IT} \sum_{itbp} \left( \frac 1T \sum_{t'}^T \sum_{rr'}^R y_{it'} y_{it}  u_{r}
                                    K_{it'rr'} \Psi_{ir'p}  u_b \right)^2 \\
    &= \frac 1{IT} \Ln \b u \Rn^2 \sum_{itp} y_{it}^2 \left( 
                \frac 1T \sum_{t'}^T \sum_{rr'}^R y_{it'}  u_{r} K_{it'rr'} \Psi_{ir'p} 
          \right)^2 \\
    &= \frac 1I \sum_i \frac 1{T^3} \sum_{t t' t''} 
                        y_{it}^2 y_{it'} y_{it''} 
                        \b u^T \bb K_{it'} \b \Psi_{i}^2 \bb K_{it''}^T \b u \\
    &\equiv \frac 1I \sum_i m_i
\end{align}
where $m_i, m_j$ are independent for $i \neq j$. Again, using Isserli's theorem \cite{isserlis1918formula} the following bounds can be calculated:
\begin{align}
    \E{m_i} \leq c' \frac {L_i^2} {\Ln \b \Gamma \Rn_2^2} \left ( R + \frac {R^3}T\right)  \quad ; \quad 
    \Var{m_i} \leq c'' \frac {L_i^4} {\Ln \b \Gamma \Rn_2^4} \left ( \frac {R^4}T + \frac {R^6} {T^2} \right)
\end{align}
where $c', c''$ are absolute constants and $L_i$ are defined in eq. (\ref{L_D_M_L_def}) and in addition, 
\begin{align}
\E{\frac 1{IT} \sum_{itbp}d_{itbp}^2 } &= \E{\frac 1I \sum_i m_i} 
    \leq c' \frac D {\Ln \b \Gamma \Rn_2^2}\left( R + \frac {R^3} {T} \right) \\
\Var{\frac 1{IT} \sum_{itbp}d_{itbp}^2  } &= \Var{\frac 1I \sum_i {m_i}} 
    \leq c'' \frac {M^2} {\Ln \b \Gamma \Rn_2^4} \left ( \frac {R^4}{IT} + \frac {R^6} {IT^2} \right)
\end{align}
where $D$ and $M$ are defined in eq. (\ref{L_D_M_L_def}). Thus, using Chebysheff inequality, it holds that
\begin{equation}
\P{\frac 1{IT}\sum_{itbp} d_{itbp} ^2 - \E{\frac 1{IT}\sum_{itbp} d_{itbp} ^2} \geq \epsilon} 
    \leq  \frac 1{IT}\left( R^4 + \frac {R^6}T \right) c'' \frac {M^2} {\Ln \b \Gamma \Rn_2^4}  \epsilon^{-2}
\end{equation}
which can be written as
\begin{equation} \label{sum_d_bounded}
    \P{\frac 1{IT}\sum_{itbp} d_{itbp}^2 \geq c''' \frac {D+M} {\Ln \b \Gamma \Rn_2^2} 
                R^2 \left(1 + \frac R {\sqrt T} \right)}  \leq \frac 1{IT}.
\end{equation}
where $c'''$ is an absolute constant that depends on $c'$ and $c''$.

\paragraph{The bound on \texorpdfstring{$\b E_2$}{Lg}}
We define the event $Q_1$ to be the event that $\frac 1{IT} \sum_{itbp} d_{itbp}^2 $ is indeed bounded by bound presented in eq. (\ref{sum_d_bounded}):
\begin{equation} \label{Q_1_event_def}
Q_1 : \quad \frac 1{IT} \sum_{itbp} d_{itbp}^2 \leq c''' R^2  \frac {D+M}{\Ln \b \Gamma \Rn_2^2} 
                    \left( 1+ \frac R{\sqrt T} \right) 
\quad ; \quad
\P{Q_1} \geq 1 - \frac 1 {IT}
\end{equation}
Under this event, the condition in eq. (\ref{IT_given_d}) gets the form
\begin{equation}
IT \geq C_2(\epsilon) R^2 B\max \left\{ 1, \frac R {\sqrt T} \right\}    
    \quad ; \quad C_2(\epsilon) \equiv \frac {4c'''}c \cdot \frac {(D+M)} {\Ln \b \Gamma \Rn_2^2\epsilon^2}
\end{equation}
and if it holds, then
\begin{equation}
\Pgiven{\Ln \b E_2\Rn_2 \geq \epsilon} {Q_1} \leq e \cdot e^{-B}.
\end{equation}
and using the union bound, it holds that
\begin{equation}
\P{\Ln \b E_2\Rn_2 \geq \epsilon} \leq \P{\overline Q_1} + \Pgiven{\Ln \b E_2\Rn_2 \geq \epsilon} {Q_1}
\leq \frac 1 {IT} + e \cdot e^{-B}
\end{equation}
\end{proof}

%%%%%%%%%%%%%%%%%%%%%%%%%%%%%%%%%%%%%%%%%%%%%%%%%%%%%%%
\subsubsection{Bound \texorpdfstring{$\b E_3$}{Lg}}
%%%%%%%%%%%%%%%%%%%%%%%%%%%%%%%%%%%%%%%%%%%%%%%%%%%%%%%

\begin{lemma}
If it holds that
\begin{equation} \label{E_3_conds}
    IT \geq C_3(\epsilon) \cdot R^2 \cdot \max \left\{\frac PT, 1\right\}
        \cdot \max \left\{ \frac {\log (IT)} T, 1\right\}
        \cdot B 
\end{equation}
where
\begin{equation} \label{C_3_def}
C_3(\epsilon) \equiv C_3' \max \left\{ 
    \frac {L}{\Ln \b \Gamma \Rn_2 \epsilon }, \frac {L^2} {\Ln \b \Gamma \Rn_2^2  \epsilon^2 } \right\}
\end{equation}
$C_3'$ is an absolute constant and $L$ is defined in eq. (\ref{L_D_M_L_def}), then
\begin{equation}
\P{\Ln \b E_3 - \E{\b E_3}\Rn_2 \geq \epsilon} \leq \frac 1 {IT} +  2 \cdot e^{-B}.
\end{equation}
\end{lemma}

\begin{proof}
Similarly to what was done to bound $\b E_2$, we first assume that $\bb K_{it}$ are constant and bound $\b E_3$ given them, and than bound $\bb K_{it}$ separately. According to the definition of the $\ell_2$ norm, it holds that
\begin{equation}
\Ln \b E_3 - \E{\b E_3} \Rn_2 
    = \max_{\b u \in \mathbb B^B_{1}} \ \b u^T \b E_3 \b u - \E{\b u^T \b E_3 \b u }
\end{equation}
where $\mathbb B^n$ is the unit-ball of $\mathbb R^n$. We again follow \cite{candes2015phase} and present a bound given $\b u$, and afterwards we use a $1/4$-Net argument to prove it for $\b u$ that depends on $\b E_3$. We notice that
\begin{align}
\b u^T \b E_3 \b u
    &= \frac1 {IT^2} \sum_{itt'} y_{it} y_{it'} 
                \left( \b u ^T \bt K_{it} \b \Psi_i \bt K_{it'}^T  \b u \right) \\
    &= \frac 1I \sum_i \Ln \frac 1T\sum_t y_{it} 
                        \left(\b \Psi_i^{1/2}\bt K_{it}^T \b u \right) \Rn^2  \\
    &= \frac 1I \sum_{ip} \left(\frac 1T \sum_t y_{it} 
                \left(\b e_p^T\b \Psi_i^{1/2}\bt K_{it}^T \b u \right) \right)^2 \\
    &\equiv \frac 1I \sum_{ip} a_{ip}^2
\end{align}
and thus
\begin{equation} \label{E_3_max_u_def}
\Ln \b E_3 - \E{\b E_3} \Rn_2 
    = \max_{\b u \in \mathbb B^{B}} \frac 1I \sum_{ip} \left( a_{ip}^2 -\E{a_{ip}^2}\right)
\end{equation}
Since $y_{it}$ are independent from $\bt K_{it}$, we see that given $\bb K_{it}$ it holds $a_{ip}$ is a sum of Gaussian scalar random variables, thus is a Gaussian random variable. Using this we notice that the element in eq. (\ref{E_3_max_u_def}) is a sum of centered squared Gaussian random variables, and thus can be bounded using Bernstein inequality. In order to use it, we first calculate the variance of $a_{ip}$:
\begin{align}
\Var{ a_{ip} }
    &= \Var{ \frac 1T \sum_{t} y_{it} 
            \left(\b e_p^T\b \Psi_i^{1/2} \bb K_{it}^T \b u \right) } \\
    &= \Var{\frac 1T \sum_{tbq} y_{it} 
            [\b \Psi^{1/2}]_{ipq} \bar K_{itbq} u_b} \\
    &\leq  \frac 1{T^2} \sum_{tbq} y_{it} ^2
            [\b \Psi^{1/2}]^2_{ipq} u_b^2 \Var{ \bar K_{itbp} } \\
    &\leq  \frac 1{T^2} \sum_{tbq} y_{it} ^2
            [\b \Psi^{1/2}]^2_{ipq} u_b^2
               && \leftarrow \text{summing some extra positive elements}\\
    &\leq \frac 1T \left( \frac 1T \sum_{t} y_{it} ^2\right)
            \Ln \b \Psi^{1/2}_{ip} \Rn^2 \Ln \b u \Rn^2 \\
    &\leq \frac 1T \left( \frac 1T \sum_{t} y_{it} ^2\right)
            \Ln \b \Psi^{1/2}_{ip} \Rn^2
                &&\leftarrow \Ln \b u \Rn=1 \text{  by definition}
\end{align}
where $\b \Psi_{ip}^{1/2}$ representes the $p$'th  row in $\b \Psi_i^{1/2}$.
According to Lemma 5.14 (sub-exponential is sub-gaussian squared) and remark 5.18 (centering lemma) in \cite{vershynin2010introduction}, we get that
\begin{align}
\Ln a_{ip}^2 - \E{a_{ip}^2} \Rn_{\psi_1} 
    \leq 2\Ln a_{ip}^2 \Rn_{\psi_1}
    \leq 4 \Ln a_{ip} \Rn_{\psi_2}^2 \leq \frac {4}T  \left( \frac 1T \sum_{t} y_{it} ^2\right)
            \Ln \b \Psi^{1/2}_{ip} \Rn^2
\end{align}
and thus, the maximal sub-exponential norm is
\begin{align}
\max_{i, p}\ \Ln a_{ip}^2 - \E{a_{ip}^2} \Rn_{\psi_1}
    &= \max_{i,p} \frac {4}T  \left( \frac 1T \sum_{t} y_{it} ^2\right)
            \Ln \b \Psi^{1/2}_{ip} \Rn^2 \\
    &= \max_i \left[ \frac {4}T  \left( \frac 1T \sum_{t} y_{it} ^2\right)
            \left (\max_p \Ln \b \Psi^{1/2}_{ip} \Rn^2 \right) \right]\\
    &\leq \max_i \left[ \frac {4}T  \left( \frac 1T \sum_{t} y_{it} ^2\right)
            \Ln \b \Psi_i \Rn_2 \right]\\
    &=  \frac {4}T  \cdot \max_i \left[  \left( \frac 1T \sum_{t} y_{it} ^2\right)
            \Ln \b \Delta_i \Rn_2 \right]\\
    & \equiv \frac {4} {T} K
\end{align}
where
\begin{equation}
K \equiv  \max_i \left[  \left( \frac 1T \sum_{t} y_{it} ^2\right)
            \Ln \b \Delta_i \Rn_2 \right].
\end{equation}
Now, we can use Bernstein inequality as it appears in Proposition 5.16 in \cite{vershynin2010introduction} to bound $\b u^T \left( \b E_3 - \E{\b E_3} \right) \b u$ in the following way:
\begin{align} \label{E_3_given_u}
\Pgiven{\b u^T \left( \b E_3 - \E{\b E_3} \right) \b u \geq \epsilon }  {\bb K_{it}}
    &\leq \Pgiven{\frac 1I \sum_{ip} \left(a_{ip}^2 - \E{a_{ip}^2}\right) \geq \epsilon } {\bb K_{it}}\\
    &\leq 2 \exp \left[ -c' \min \left\{ 
            \frac {I T^2 \cdot \epsilon^2} {16 c^2 P K^2},
            \frac {I T \cdot \epsilon} {4cK}\right\} \right] 
\end{align}
where $c'$ is an absolute constant that originates from Bernstein inequality. 
To give a bound for the extremal $\b u$, we use an $1/4$-net in the unit ball $\mathbb B^B$, and using lemma 5.4 in \cite{vershynin2010introduction} we get that:
\begin{equation}
    \max_{\b u  \in \mathbb B^{B}}\  \b u^T \left( \b E_3 - \E{\b E_3} \right) \b u
    \leq 2\max_{\b u \in \mathcal N_{1/4}}\ \b u^T \left( \b E_3 - \E{\b E_3} \right) \b u
\end{equation}
Since eq. (\ref{E_3_given_u}) holds for every independent $\b u$, we can use the union bound to get
\begin{align}
\Pgiven{2\max_{\bt y \in \mathcal N_{1/4}} \b u^T \left( \b E_3 - \E{\b E_3} \right) \b u \leq \epsilon} {\bb K_{it}}
    &\geq 1- 2 \cdot 9^B \cdot \exp \left[ -c' \min \left\{ 
            \frac {I T^2 \cdot \epsilon^2} {16 c^2 P K^2},
            \frac {I T \cdot \epsilon} {4cK}\right\} \right]  \nonumber \\
    &\geq 1- 2 \cdot \exp \left[3B -c' \min \left\{ 
            \frac {I T^2 \cdot \epsilon^2} {16 c^2 P K^2},
            \frac {I T \cdot \epsilon} {4cK}\right\} \right] \label{P_E_3_given_K_no_conds}
\end{align}
and if we require that
\begin{equation} \label{E_3_conds_given_K}
IT \geq \frac 4 {c'} \max \left\{ \frac PT, 1\right\} 
    \cdot \max \left\{\frac {4cK} \epsilon,  \left(\frac {4cK} \epsilon \right)^2
            \right\} 
    \cdot B
\end{equation}
it holds that
\begin{equation}
3B -c' \min \left\{ 
            \frac {I T^2 \cdot \epsilon^2} {16 c^2 P K^2},
            \frac {I T \cdot \epsilon} {4cK}\right\} \leq -B
\end{equation}
and if we put this result in eq. (\ref{P_E_3_given_K_no_conds}), we get that
\begin{equation}
\Pgiven{\Ln \b E_3 - \E{\b E_3} \Rn_2 \geq \epsilon} {\bb K_{it}} \leq 2 e^{-B}.
\end{equation}

\paragraph{Bounding \texorpdfstring{$K$}{Lg}:}
Next, we show that with great probability, $K$ from eq. (\ref{E_3_conds_given_K}) is bounded by at most logarithmic factor. Since $K$ depends solely on $\bb K_{it}$ and thus is independent of $\bt K$ and $\b u$ we can bound it separately. We follow $K$ definition and denote
\begin{equation}
K \equiv \max_i\ \frac 1T \sum_t  y_{it}^2 \Ln \b \Delta_i \Rn_2 \equiv \max_i b_i
\end{equation}
and we show that $b_i$ is subexponential, and thus the maximal $b_i$ is far from its expectation by at most a logarithmic factor in $I$.  It can be seen that the expectation of $b_i$ satisfy
\begin{gather}
\E{b_i} = \E{\frac 1T \sum_t y_{it}^2\Ln \b \Delta_i \Rn_2} 
        = \Ln \b \Delta_i \Rn_2 \E{y_{it}^2} \\
\Longrightarrow \quad
b_i - \E{b_i} = \frac 1T\sum_t \Ln \b \Delta_i \Rn_2 \left( y_{it}^2 - \E{y_{it}^2} \right)
\end{gather}
an we notice that $\Ln \b \Delta_i \Rn_2 (y_{it}^2 - \E{y_{it}^2})$ is centered squared Gaussian random variable, and thus is sub-exponential. To calculate its sub-exponential norm we first calculate the variance of $y_{it}$
\begin{align} \label{yit_subgauss_norm}
\Var{ y_{it} }
    &= \Var{\sum_{r, r', r''} W^{(1)}_{rr'} \bar K_{r' r''} V^{(1)}_{ir'' r} }
        && \leftarrow \text{definition of $y_{it}$} \\
    &\leq \sum_{r, r', r''} \left(W^{(1)}_{rr'} V^{(1)}_{ir'' r}\right)^2  \\
    &\leq R \Ln \b W^{(1)} \Rn_2^2 \Ln \b V_i^{(1)} \Rn_2^2 \\ 
    &= R \Ln \b \Gamma^{1/2} \b W \Rn_2^2 \Ln \b \Delta_i^{1/2} \b V_i \Rn_2^2 
        && \leftarrow \text{by eq. (\ref{WV_1_def})}
\end{align}
and since sub-exponential is sub-Gaussian squared and using the centering lemma \cite{vershynin2010introduction}, we get that
\begin{align}
\Ln \Ln \b \Delta_i \Rn_2 \left( y_{it}^2 - \E{y_{it}^2} \right)\Rn_{\psi_1} 
    &\leq \Ln \b \Delta_i \Rn_2  \Ln y_{it}^2 - \E{y_{it}^2}\Rn_{\psi_1}  
            && \\
    &\leq 2 \Ln \b \Delta_i \Rn_2 \Ln y_{it}^2\Rn_{\psi_1} 
            &&\leftarrow \text{centering lemma} \\
    &\leq 4\Ln \b \Delta_i \Rn_2  \Ln y_{it}\Rn_{\psi_2}^2 
            &&\leftarrow \text{subexp is subgauss squared} \nonumber\\
    &\leq 4R \Ln \b W \Rn_2^2 \Ln \b V_i \Rn_2^2 
                \Ln \b \Gamma \Rn_2 \Ln \b \Delta_i\Rn_2^2 
            &&\leftarrow \text{by eq. (\ref{yit_subgauss_norm})} \\
    &\leq \frac {4RL} {\Ln \b \Gamma \Rn_2}
\end{align}
where $L$ is as defined in eq. (\ref{L_D_M_L_def}). Using Bernstein inequality, we can see that
\begin{align}
\forall i \quad
\P{ b_i \geq \E{b_i} + \delta}
    &=\P{\frac 1T \sum_t \Ln \b \Delta_i \Rn_2  \left( y_{it}^2 - \E{y_{it}^2} \right) \geq \delta}  \\
    &\leq 2 \exp\left[ -c'T \min \left\{  
            \frac {\Ln \b \Gamma \Rn_2^2\delta^2} {16R^2L^2},  
            \frac {\Ln \b \Gamma \Rn_2\delta} {4RL}
    \right\}\right] 
\end{align}
where $c'$ is an absolute constant that originates from Bernstein inequality. Using the union bound, it holds that
\begin{align}
\P{K  \geq \E{b_i} + \delta} 
    &= \P{ \max_i b_i \geq \E{b_i}+ \delta} \\
    &\leq 2I \exp\left[ -c'T \min \left\{ 
            \frac {\Ln \b \Gamma \Rn_2^2\delta^2} {16R^2L^2},  
            \frac {\Ln \b \Gamma \Rn_2\delta} {4RL}
    \right\}\right]
\end{align}
and by demanding that it will hold with probability of $1 - \frac 1 {IT}$, we get:
\begin{gather} \label{E_3_delta_cond_complex}
2I \exp\left[ -c'T \min \left\{ 
            \frac {\Ln \b \Gamma \Rn_2^2\delta^2} {16R^2L^2},  
            \frac {\Ln \b \Gamma \Rn_2\delta} {4RL}
    \right\}\right] \leq \frac 1 {IT} \\
\Longrightarrow\quad
c' T \min \left\{             
            \frac {\Ln \b \Gamma \Rn_2^2\delta^2} {16R^2L^2},  
            \frac {\Ln \b \Gamma \Rn_2\delta} {4RL}
\right\} 
    \geq  \log(2 I^2 T)
\end{gather}
In this setting, we try to bound from above $\delta$, which is the distance between $K$ and its expectation with the above probability. Thus, we can demand that
\begin{equation}  \label{E_3_delta_simplify_cond}
\delta \geq \frac {4 R L } {\Ln \b \Gamma \Rn_2}
\end{equation}
to simplify the condition in eq. (\ref{E_3_delta_cond_complex}) and get
\begin{gather} \label{E_3_delta_fixed_cond}
c' T \frac {\Ln \b \Gamma \Rn_2 \delta} {4RL}  \geq  \log(2 I^2 T) 
\quad \Longrightarrow \quad
\delta \geq \frac {4\log(2I^2 T)} {c' T} \cdot \frac {RL}{\Ln \b \Gamma \Rn_2}
\end{gather}
and by combining the conditions in $\delta$ in eq. (\ref{E_3_delta_simplify_cond}) and eq. (\ref{E_3_delta_fixed_cond}), we get that:
\begin{equation}
\delta \geq \frac {RL}{\Ln \b \Gamma \Rn_2} \left( c''\frac {\log(I T)} T + 4 \right)    
\end{equation}
where $c''$ is some absolute constant depending on $c'$, and thus
\begin{equation} \label{E_3_K_bound_with_expect}
\P{K \geq \E{b_i} + \frac {RL}{\Ln \b \Gamma \Rn_2}\left( c'' \frac {\log(IT)} T + 4\right) } \leq \frac 1 {IT}.
\end{equation}

Next, in order to bound $\E{b_{it}}$ we notice that, 
\begin{align}
\E{b_i}  =  \Ln \b \Delta_i \Rn_2 \E{y_{it}^2}
    = \Ln \b \Delta_i \Rn_2 R \Ln \b \Gamma^{1/2} \b W \Rn_2^2 \Ln \b \Delta_i^{1/2} \b V_i \Rn_2^2  
    \leq \frac {RL}{\Ln \b \Gamma \Rn_2}
\end{align}
and by putting it back in eq. (\ref{E_3_K_bound_with_expect}), we can bound $K$ by the following
\begin{equation} \label{K_bound}
\P{K \geq \frac {RL}{\Ln \b \Gamma \Rn_2} \left( c'' \frac {\log(IT)} T + 5\right) } \leq \frac 1 {IT}
\end{equation}

\paragraph{The bound on \texorpdfstring{$\b E_3$}{Lg}:}
We define the event $Q_2$ to be the event that $K$ is indeed bounded by the bound presented in eq. (\ref{K_bound}):
\begin{equation} \label{Q_2_event_def}
    Q_2 : \quad K \leq \frac {RL}{\Ln \b \Gamma \Rn_2} \left( c'' \frac {\log(IT)} T + 5\right)
    \quad ; \quad
    \P{Q_2} \geq 1 - \frac 1{IT}
\end{equation}
under this event, the conditions in eq. (\ref{E_3_conds_given_K}) get the following form
\begin{equation}
    IT \geq C_3(\epsilon) \cdot R^2 \cdot \max \left\{\frac PT, 1\right\}
        \cdot \max \left\{ \frac {\log (IT)} T, 1\right\}
        \cdot B 
\end{equation}
where 
\begin{equation}
    C_3(\epsilon) \equiv C_3' \max \left\{ \frac L{\Ln \b \Gamma \Rn_2 \epsilon}, 
                                          \frac {L^2}{ \Ln \b \Gamma \Rn_2^2 \epsilon^2} \right\}
\end{equation}
and $C_3'$ is some constant depending on $c, c'$ and $c''$. If these condition hold, it holds that
\begin{equation}
\Pgiven{\Ln \b E_3 - \E{\b E_3} \Rn_2 \geq \epsilon} {Q_2} \leq 2 e^{-B} .
\end{equation}
and totally, using the union bound, it holds that
\begin{equation}
    \P{\Ln \b E_3 - \E{\b E_3} \Rn_2 \geq \epsilon} \leq 
        \P{\overline Q_2} + \Pgiven{\Ln \b E_3 - \E{\b E_3} \Rn_2 \geq \epsilon} {Q_2} \leq \frac 1{IT} + 2e^{-B}
\end{equation}
\end{proof}

%%%%%%%%%%%%%%%%%%%%%%%%%%%%%%%%%%%%%%%%%%%%%%%%%%%%%%%
\subsubsection{Union Bound}
%%%%%%%%%%%%%%%%%%%%%%%%%%%%%%%%%%%%%%%%%%%%%%%%%%%%%%%
By combining eq. (\ref{standart_A}, \ref{split_A}) and the union bound we get that
\begin{multline}
\P{\Ln \b A - \bh A\Rn_2 \geq \gamma} 
    \leq \P{\Ln \b E_1 - \E{\b E_1}\Rn_2 \geq \frac \gamma {3 \Ln \b \Gamma \Rn_2^2}}
        + \P{\Ln \b E_2 - \E{\b E_2}\Rn_2 \geq \frac \gamma {3 \Ln \b \Gamma \Rn_2^2}}  \\
        + \P{\Ln \b E_3 - \E{\b E_3}\Rn_2 \geq \frac \gamma {3 \Ln \b \Gamma \Rn_2^2}}
\end{multline}
and, if the conditions on $I$ and $T$ in eq. (\ref{E_1_cond}, \ref{E_2_cond}, \ref{E_3_conds}) hold, we get that  $\b E_1, \b E_2$ and $\b E_3$ are close to their expectations, i.e.
\begin{equation} \label{P_A_dist_with_it}
\P{\Ln \b A - \bh A\Rn_2 \geq \gamma} 
    \leq \frac 1{3B} 
        + \frac 2{IT}
        + e \cdot e^{-B}
        + 2 e ^{-B}
\end{equation}
and if we further require that:
\begin{equation} \label{extra_IT_cond}
        IT \geq 3 B
\end{equation}
then eq. (\ref{P_A_dist_with_it}) gets the form
\begin{equation}
\P{\Ln \b A - \bh A\Rn_2 \geq \gamma} 
    \leq \frac 1{B} 
        + e \cdot e^{-B}
        + 2 e ^{-B}
\end{equation}
In addition, in order for the conditions in eq. (\ref{E_1_cond}, \ref{E_2_cond}, \ref{E_3_conds}, \ref{extra_IT_cond}) to hold, we require that:
\begin{equation} \label{total_conds}
    IT \geq C(\gamma) \cdot R^3  \cdot
                     \max\left\{\frac RT, 1\right\}^2 \cdot 
                     \max \left \{\frac {\log(IT)} T, 1 \right\}^2 \cdot
                     \max\left\{\frac PT, 1\right\} \cdot
                     B 
\end{equation}
where
\begin{align}
C(\gamma) &= \underbrace{C_1\left( \frac \gamma {3\Ln \b \Gamma \Rn_2^2}\right)}
                        _{\text{defined in eq. (\ref{C_1_def})}}
                + \underbrace{C_2\left( \frac \gamma {3\Ln \b \Gamma \Rn_2^2}\right)}
                        _{\text{defined in eq. (\ref{C_2_def})}}
                + \underbrace{C_3\left( \frac \gamma {3\Ln \b \Gamma \Rn_2^2}\right)}
                        _{\text{defined in eq. (\ref{C_3_def})}}
                + \underbrace{3}
                        _{\text{by eq. (\ref{extra_IT_cond})}} \\
    &= c_1 D \gamma^{-2} + c_2 M \gamma^{-2} + c_3 \max \left\{L \epsilon^{-1}, L^2 \epsilon^{-2} \right\}
\end{align}
for $c_1, c_2, c_3$ absolute constants that can be caluclated from $C_1', C_2', C_3'$, and totally, if this condition holds, it holds that
\begin{equation}
\P{\Ln \b A - \bh A\Rn_2 \geq \gamma} 
    \leq \frac 1B
        + 5 \cdot e^{-B}.
\end{equation}

\qed

%%%%%%%%%%%%%%%%%%%%%%%%%%%%%%%%%%%%%%%%%%%%%%%%%%%%%%%%%%%%%%%%%%%%%%%%%%%%%%%5
%% Proof of Lambda Concentration
%%%%%%%%%%%%%%%%%%%%%%%%%%%%%%%%%%%%%%%%%%%%%%%%%

\subsection{Proof of Lemma \ref{lambda_close_lemma}}
By the $\ell_2$ norm definition, it holds that:
\begin{equation} \label{gamma_l2_diff_def}
\Ln \bh \Gamma - \E{\bh \Gamma} \Rn_2 
    = \max_{\b u \in \mathbb B^B} \left| \b u^T \bh \Gamma \b u - \E{\b u^T \bh \Gamma \b u }\right|
\end{equation}
Similarly to lemma 5.50 in \cite{vershynin2010introduction}, we compute the bound given a constant independent $\b u$ and then use an $\epsilon$-Net together with union bound to complete the proof.
We first denote:
\begin{equation}
\b X_{it} \equiv \b \Gamma^{1/2} \b R_{it} \b \Delta_i^{1/2} 
\quad ; \quad
\b R_{it} \sim \mathcal N(\b 0_{B\times P}, \b I_B \otimes \b I_{P})
\end{equation}
and using SVD decomposition, we denote:
\begin{equation}
\b \Delta_i = \b B_i \b \Sigma_i \b B_i^T
\end{equation}
where $\b B_i \in \mathbb R^{B\times B}$ are orthogonal matrices and 
$\b \Sigma_i \equiv \text{diag}(\lambda_1^{(i)}, \lambda_2^{(i)}, ..., \lambda_P^{(i)})$ 
is a diagonal matrix. To simplify the notation, we further denote
\begin{equation}
\bt u \equiv \b \Gamma^{1/2} \b u.
\end{equation}
Since $\b R_{it}$ are matrices of i.i.d. Gaussian elements, we can use the spherically-symmetric property to define:
\begin{equation}
\b R_{it} = \b K_{it} \b B^T 
\quad ; \quad 
\b K_{it} \sim \mathcal N(\b 0_{B\times P}, \b I_B \otimes \b I_{P})
\end{equation}

Using these, when looking at the elements in eq. (\ref{gamma_l2_diff_def}), we get:
\begin{align}
\b u^T \bh \Gamma \b u 
    &= \frac 1{ITP} \sum_{it} \b u^T \b X_{it} \b X_{it}^T \b u \\
    &= \frac 1{ITP} \sum_{it} \Ln \b X_{it}^T \b u \Rn^2 \\
    &= \frac 1{ITP} \sum_{it} \Ln \b B_i \b \Sigma_i^{1/2} \b B_i^T \b B_i \b K_{it}^T \bt u \Rn^2 \\
    &= \frac 1{ITP} \sum_{it} \Ln \b \Sigma_i^{1/2} \b K_{it}^T \bt u \Rn^2 \\
    &= \frac 1{ITP} \sum_{itp} \lambda_p^{(i)} \left(\b k_{it:p}^T \bt u \right)^2
\end{align}
where $\b k_{it:p}\in \mathbb R^B$ is the $p$'th column  of $\b K_{it}$.  Thus, it holds that
\begin{align}
\P{ \left| \b u^T \bh \Gamma \b u - \E{\b u^T \bh \Gamma \b u} \right| \geq \epsilon}
    &= \P{ \left| \sum_{itp} \frac{\lambda_{p}^{(i)}} {ITP}  \Big( 
            (\b k_{it:p}^T \bt u)^2  - \E{(\b k_{it:p}^T \bt u)^2}\Big)
        \right| \geq \epsilon } \\
    &\equiv \P{ \left| \sum_{itp} \frac{\lambda_{p}^{(i)}} {ITP}  a_{itp}
        \right| \geq \epsilon }
\end{align}
In order to bound this, we use Bernstein inequality. According to Lemma 5.9 (sub-Gaussian rotation invariance), Lemma 5.14 (sub-exponential is sub-gaussian squared) and remark 5.18 (centering Lemma) in \cite{vershynin2010introduction}, it can be seen that $a_{itp}$ are all scalar sub-exponential random variables with the following sub-exponential norm:
\begin{gather}
\forall itp \quad \Ln (\b k_{it:p}^T \bt u)^2\Rn_{\psi_1} 
    \leq 2 \Ln \b k_{it:p}^T \bt u \Rn_{\psi_1}^2 
    \leq 2c \Ln \bt u \Rn^2 \leq 2c \Ln \b \Gamma\Rn_2 \\ 
\Longrightarrow \quad 
\Ln a_{itp} \Rn_{\psi_1} = \Ln (\b k_{it:p}^T \bt u)^2  - \E{(\b k_{it:p}^T \bt u)^2} \Rn_{\psi_1} 
    \leq 4c \Ln \b \Gamma\Rn_2
\end{gather}
for some constant $c$. To use Bernstein inequality, we first compute the following elements:
\begin{align}
&\sum_{itp} \left( \frac{\lambda_{p}^{(i)}} {ITP} \right)^2
    \leq \frac 1{ITP} \cdot \frac 1I \sum_i \Ln \b \Delta_i \Rn_2 ^2  
    \equiv \frac 1{ITP} \bar K \\
&\max_{itp} \left( \frac{\lambda_{p}^{(i)}} {ITP} \right)
    = \frac 1{ITP} \cdot \max_i \Ln \b \Delta_i \Rn_2
    \equiv \frac 1{ITP} K_{max}
\end{align}
and using Bernstein inequality, we get that
\begin{align} \label{Gamma_bound_given_u}
\P{ \left| \b u^T \bh \Gamma \b u - \E{\b u^T \bh \Gamma \b u} \right| \geq \epsilon}
    &= \P{ \left| \sum_{itp} \frac{\lambda_{p}^{(i)}} {ITP}  a_{itp}
        \right| \geq \epsilon } \\
    % &\leq 2\exp \left[ -c' ITP \min \left\{ 
    %         \frac {\epsilon^2} {\Ln \b \Gamma \Rn_2^2  \frac 1I \sum_i \Ln \b \Delta_i \Rn_2^2 },
    %         \frac {\epsilon} {\Ln \b \Gamma \Rn_2 \max_i \Ln \b \Delta_i \Rn_2}
    %     \right\} \right] \\
    &= 2\exp \left[ -c' ITP \min \left\{ 
            \frac {\epsilon^2} {\Ln \b \Gamma\Rn_2^2 \bar K},
            \frac {\epsilon} {\Ln \b \Gamma\Rn_2 K_{max}}
        \right\} \right]
\end{align}
where $c'$ is an absolute constant that depends both on the constant in Bernstein inequality and on $c, c^2$. To give a bound for the extremal $\b u$, we use an $1/4$-net in the unit ball $\mathbb B^B$. According to lemma 5.2 in \cite{vershynin2010introduction}, to cover the unit ball in this vector space, such net needs a total of
\begin{equation}
     | \mathcal N_{1/4} | = 9^{B}
\end{equation}
points, and using lemma 5.4 in \cite{vershynin2010introduction} we get that
\begin{equation}
\Ln \bh \Gamma - \E{\bh \Gamma} \Rn_2 
\leq 2 \max_{\b u \in \mathcal N(B)} \left( \b u^T \bh \Gamma \b u 
        - \E {\b u ^T \bh \Gamma \b u} \right).
\end{equation} 
Since eq. (\ref{Gamma_bound_given_u}) holds for every constant $\b u$, we can use the union bound to get
\begin{align} \label{Gamma_bound_with_ITP}
\P{\Ln \bh \Gamma - \E{\bh \Gamma} \Rn_2  \geq \epsilon}
    &= \P{2 \max_{\b u \in \mathcal N(B)} \left( \b u^T \bh \Gamma \b u 
        - \E {\b u ^T \bh \Gamma \b u} \right)  \geq \epsilon} \\
    &\leq 9^B \cdot 2\exp \left[ -c' ITP \min \left\{ 
            \frac {\epsilon^2} {\Ln \b \Gamma\Rn_2^2 \bar K},
            \frac {\epsilon} {\Ln \b \Gamma\Rn_2 K_{max}}
        \right\} \right] \\
    &\leq 2\exp \left[ 3B -c' ITP \min \left\{ 
            \frac {\epsilon^2} {\Ln \b \Gamma\Rn_2^2 \bar K},
            \frac {\epsilon} {\Ln \b \Gamma\Rn_2 K_{max}}
        \right\} \right]
\end{align}
thus, if we require
\begin{equation}
ITP \geq \frac 4 {c'} \max\left\{ \Ln \b \Gamma \Rn_2^2 \bar K \epsilon^{-2}, 
                                  \Ln \b \Gamma \Rn_2 K_{max}\epsilon^{-1} \right\} B
\end{equation}
it holds that
\begin{equation}
3B - c' ITP \min \left\{ 
            \frac {\epsilon^2} {\Ln \b \Gamma\Rn_2^2 \bar K},
            \frac {\epsilon} {\Ln \b \Gamma\Rn_2 K_{max}}
        \right\} \leq B \\
\end{equation}
and if we put this result in eq. (\ref{Gamma_bound_with_ITP}), we get
\begin{equation}
    \P{ \Ln \b \Gamma - \bh \Gamma \Rn_2 \geq \epsilon } \leq 2 e^{-B}
\end{equation}

\qed

\subsection{Expansion of the constant \texorpdfstring{$k$}{Lg}} \label{k_expansion_section}
As explained before, the constant $k$ in eq. (\ref{theorem_IT_cond}) depends solely on the matrices $\b \Gamma, \b \Delta^{(i)}, \b W, \b V_i$. Though only a constant, a closer look on the expansion of $k$ can shed some light on the relation between the different tasks. For simpler notation, we start by defining the following task-divergence coefficients:
\begin{equation} \label{div_parameters_def} \begin{split}
\eta \equiv \frac {\frac 1I \sum_i L_i}
                            {\lambda_{\min}\left(\b \Gamma \b W \b Q \b W^T \b \Gamma \right)}
&,\quad
\alpha \equiv \frac {\frac 1I \sum_i L_i^2} 
                            {\lambda_{\min}\left(\b \Gamma \b W \b Q \b W^T \b \Gamma \right)^2} 
, \\
\mu \equiv \frac {\sqrt {\frac 1I \sum_i L_i^4}} 
                            {\lambda_{\min}\left(\b \Gamma \b W \b Q \b W^T \b \Gamma \right)^2}
&,\quad
\nu \equiv \frac {\max_i L_i} 
                            {\lambda_{\min}\left(\b \Gamma \b W \b Q \b W^T \b \Gamma \right)}
,\\
\psi \equiv \frac {\frac 1I \sum_i \Ln \b \Delta_i \Rn_2^2} 
                            {\left(\frac 1{IP} \sum_i \tr{ \b \Delta_i} \right)^2} = \bar K
&,\quad
\chi \equiv \frac {\max_i \Ln \b \Delta_i \Rn_2} 
                            {\frac 1{IP} \sum_i \tr{\b \Delta_i}} = K_{\max} \\
\kappa_{\b \Gamma} &\equiv \frac {\Ln \b \Gamma \Rn_2} {\lambda_{\min} (\b \Gamma) }
\end{split} \end{equation}
where $L_i$ is defined in eq. (\ref{L_D_M_L_def}) and $\bar K, K_{\max}$ are defined in eq. (\ref{K_bar_max_def}).

The expansion of $k$ can be seen when looking more closely at the conditions for eq. (\ref{prob_of_conditions}) to hold. By looking at lemma \ref{A_expectation} and 
lemma \ref{lambda_close_lemma} it can be seen that in order for (\ref{prob_of_conditions}) to hold, the conditions on $IT$ are
\begin{align} \label{IT_conds_intermidate} \begin{split}
IT &\geq \frac 1P K_1(\gamma)B \\
IT &\geq \frac 1P K_2(\gamma) \cdot \max \left\{\frac {R^{1.5}}T, \frac {R^3P} {T^2} \right\}
        \cdot B \\
IT &\geq K_3(\gamma) \cdot R^4 \cdot 
     \max\left\{\frac RT, 1\right\}^2 \cdot 
     \max \left \{\frac {\log(IT)} T, 1 \right\}^2 \cdot 
     \max\left\{\frac PT, 1\right\} \cdot B 
\end{split} \end{align}
where $K_1, K_2$ and $K_3$ are defined as follows:
\begin{align} \label{K_1_K_2_K_3_def} \begin{split}
    K_1(\gamma) &= 4 D' \max \left\{
                            \kappa_{\b \Gamma} \chi,  \ 
                            \kappa_{\b \Gamma}^2 \psi \right\} \\
    K_2(\gamma) &= D' \max \left\{
                            \kappa_{\b \Gamma} \eta \chi \gamma^{-1},  \ 
                            \kappa_{\b \Gamma}^2 \eta^2 \psi \gamma^{-2} \right\}  \\
    K_3(\gamma) &= c_1\alpha\gamma^{-2} + c_2 \mu \gamma ^{-2} 
        + c_3 \max \left\{\nu \gamma^{-1}, \nu^2 \gamma^{-2} \right\}
\end{split}\end{align}
for some gloal numerical constants $c_1, c_2, c_3$ and $D'$ as defined in eq. (\ref{A_expecation_C_def}) and (\ref{lambda_close_lemma_D_def}).

We recall that $\epsilon \equiv c \cdot \gamma$ for a constant $c$ depending on $\kappa_{\b \Gamma}$ and satisfies $c > 1$. Since $\epsilon$ is used to bound $\dist{\b W, \bh W}$ and thus is only relevant in the range $[0,1]$, we get that $0 \leq \gamma \leq 1$ and thus, the following condition is sufficient for the three conditions in eq. (\ref{IT_conds_intermidate}) to hold
\begin{equation}\label{IT_conds_combined}
    IT \geq k' \cdot \gamma^{-2} \cdot R^4 \cdot 
    \max\left\{\frac RT, 1\right\}^2 \cdot 
    \max \left \{\frac {\log(IT)} T, 1 \right\}^2 \cdot
    \max\left\{\frac PT, 1\right\} \cdot B
\end{equation}
where $k'$, which is also referred in eq. (\ref{theorem_IT_cond_with_k_tag}), is defined to be:
\begin{gather}
    k' = c_1\alpha + c_2\mu + c_3 \max \left\{\nu, \nu^2 \right\} + 5 D' \max \left\{
                            \kappa_{\b \Gamma} \eta \chi,  \ 
                            \kappa_{\b \Gamma}^2 \eta^2 \psi \right\}
\end{gather}
and finally, as defined before, $k \equiv c^2 \cdot k'$.

\section*{Appendix B - Extra synthetic experiments}

The below graphs show the results of synthetic numerical experiments for the dependency between the \emph{CMR algorithm} probability of recovery of $\bh W$ with the parameters $I$ , $T$ and $B$. For simplicity, we use $R=1$, $P=1$ and use $i.i.d.$ samples, i.e. $\b \Gamma = \b I, \b \Delta^{(i)}=\b I$.
Each pixel in the graphs presents the observed probability for $\dist{\bh W, \b W} < 0.25$ as estimated by 150 different iterations of the \emph{CMR algorithm} on different random $\b X_{it}$, $\b W$ and $\b V_i$.  This two graphs show seemingly linear dependency between $B$ and $I,T$, which indicates that the bound presented in theorem 1 is indeed tight.

\begin{figure*}[h!] \label{CMR_synth}
\centering
\begin{tabular}{cc}
\includegraphics[height=0.39\textwidth]{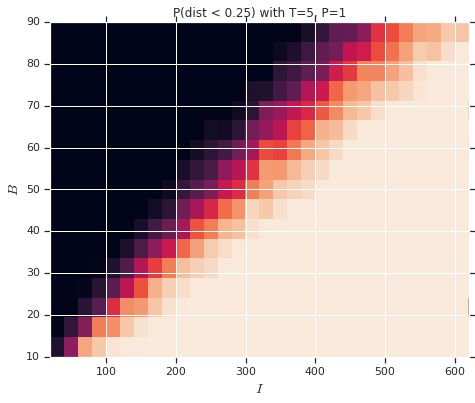} &
\includegraphics[height=0.39\textwidth]{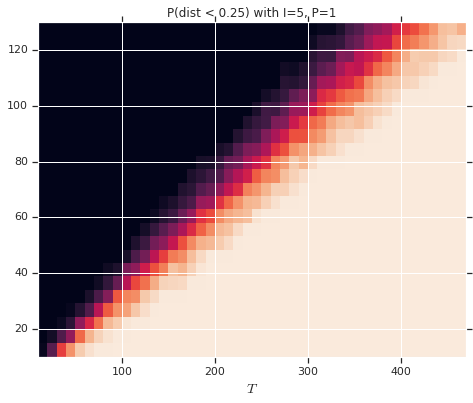}
\end{tabular}
\vspace{-0.1in}
\caption{Observed probability for recovery of the true shared mechanism $\b W$ as a function of the dimention of $\b W$ $B$ (y-axis), the number of tasks $I$ (x-axis, left) and the number of samples per task $T$ (x-axis, right)}
\label{success_graphs}
\end{figure*}

\end{document}